%% file: main.tex
\documentclass[lettersize,journal]{IEEEtran}
\usepackage{amsmath,amsfonts}
\usepackage{algorithmic}
\usepackage{algorithm}
\usepackage{array}
\usepackage[caption=false,font=normalsize,labelfont=sf,textfont=sf]{subfig}
\usepackage{textcomp}
\usepackage{stfloats}
\usepackage{url}
\usepackage{verbatim}
\usepackage{graphicx}
\usepackage{cite}
\usepackage{multirow}  
\usepackage{multicol} 
\usepackage{booktabs}
\usepackage{amsfonts}  
\usepackage{overpic}
\usepackage{enumitem} 
\usepackage{xcolor}
\usepackage{colortbl}
\usepackage{tikz}
\usepackage{pgfplots}
\usetikzlibrary{plotmarks,patterns,matrix,decorations.pathreplacing}
\pgfplotsset{compat=newest}

\usepackage[pagebackref,breaklinks,colorlinks]{hyperref}
\usepackage{xspace}

\newcommand{\Fig}[1]{Fig.~\ref{#1}}
\newcommand{\Table}[1]{Table~\ref{tab:#1}}
\usepackage{color}
\definecolor{turquoise}{cmyk}{0.65,0,0.1,0.3}
\definecolor{purple}{rgb}{0.65,0,0.65}
\definecolor{dark_green}{rgb}{0, 0.5, 0}
\definecolor{orange}{rgb}{0.8, 0.6, 0.2}
\definecolor{darkred}{rgb}{0.6, 0.1, 0.05}
\definecolor{blueish}{rgb}{0.0, 0.3, 0.6}
\definecolor{light_gray}{rgb}{0.7, 0.7, .7}
\definecolor{pink}{rgb}{1, 0, 1}
\definecolor{greyblue}{rgb}{0.25, 0.25, 1}
\definecolor{green}{RGB}{0,135,54}
\definecolor{blue}{RGB}{87, 211, 219}
\definecolor{red}{RGB}{219, 95, 87}

\makeatletter
\DeclareRobustCommand\onedot{\futurelet\@let@token\@onedot}
\def\@onedot{\ifx\@let@token.\else.\null\fi\xspace}

\def\ie{\emph{i.e}\onedot} 
 
\def\etc{\emph{etc}\onedot}

\def\etal{\emph{et al}\onedot}
\makeatother

\makeatletter
\g@addto@macro{\endtabular}{\rowfont{}}
\makeatother
\newcommand{\rowfonttype}{}
\newcommand{\rowfont}[1]{
   \gdef\rowfonttype{#1}#1%
}
\newcolumntype{C}{>{\rowfonttype}c}

\begin{document}

\title{Scene-Adaptive Attention Network for Crowd Counting\thanks{Xing Wei, Yuanrui Kang, Jihao Yang, Yunfeng Qiu, and Yihong Gong are with the school of software engineering, Xi'an Jiaotong University. Dahu Shi and Wenming Tan are with the Hikvision Research Institute.}}

\author{Xing Wei, Yuanrui Kang, Jihao Yang, Yunfeng Qiu, Dahu Shi, Wenming Tan, Yihong Gong
}


\maketitle

\input{sec/0_abstract}

\input{sec/1_introduction}

\input{sec/2_related_work}
\input{sec/3_method}
\input{sec/4_experiments}

\input{sec/5_conclusion}

 







{
    \bibliographystyle{IEEEtran}
    \bibliography{main}
}


 




\vfill

\end{document}

%% file: sec/0_abstract.tex
\begin{abstract}
	
	In recent years, significant progress has been made on the research of crowd counting. However, as the challenging scale variations and complex scenes existed in crowds, neither traditional convolution networks nor near recent Transformer architectures with fixed-size attention could handle the task well. 
	To address this problem, this paper proposes a scene-adaptive attention network, termed SAANet.
	First of all, we design a deformable attention in-built Transformer backbone, which learns adaptive feature representations with deformable sampling locations and dynamic attention weights.
	Then we propose the multi-level feature fusion and count-attentive feature enhancement modules further, to strengthen feature representation under the global image context.
	The learned representations could attend to the foreground and are adaptive to different scales of crowds.
	We conduct extensive experiments on four challenging crowd counting benchmarks, demonstrating that our method achieves state-of-the-art performance. 
	Especially, our method currently ranks No.1 on the public leaderboard of the NWPU-Crowd benchmark.
	We hope our method could be a strong baseline to support future research in crowd counting.
	The source code will be released to the community.
	
\end{abstract}


%% file: sec/1_introduction.tex
\section{Introduction}
\label{sec:intro}

The goal of crowd counting is to estimate the number of person instances within a scene that could be very crowded. Crowd counting has attracted more and more attention recently since its important applications in public security~\cite{wang2018generative}, traffic monitoring~\cite{zhang2017fcn,li2018csrnet}, agriculture~\cite{aich2017leaf,liu2018robust}, biomedicine~\cite{lempitsky2010learning}, and other object counting tasks~\cite{marsden2018people}.

\begin{figure}
  \centering
  \def\arraystretch{1.2}
  \setlength{\tabcolsep}{2pt}
  \begin{tabular}{CCC}
    \includegraphics[width=0.31\linewidth]{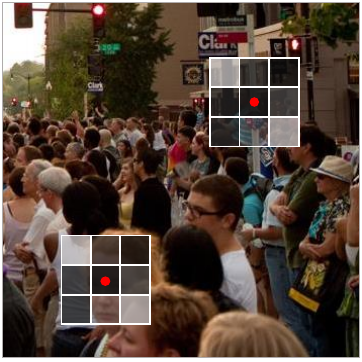}&
    \includegraphics[width=0.31\linewidth]{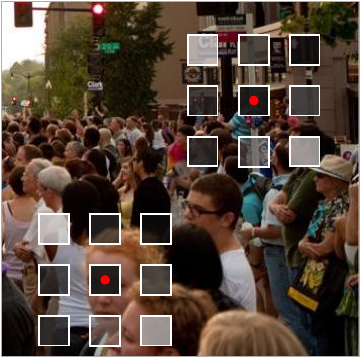}&
    \includegraphics[width=0.31\linewidth]{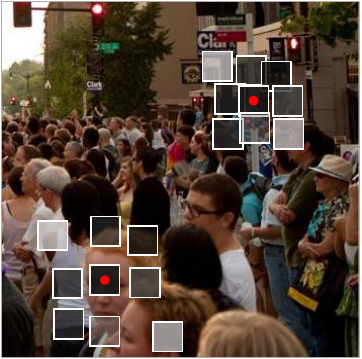}\\
    \rowfont{\scriptsize}
    (a) Regular Conv. & (b) Dilated Conv. & (c) Deformable Conv.\\ 
    \includegraphics[width=0.31\linewidth]{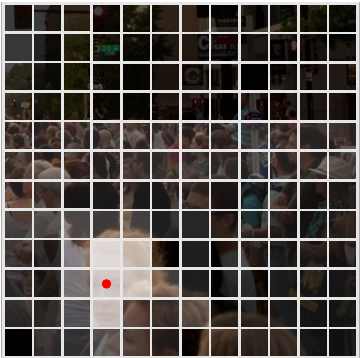}& 
    \includegraphics[width=0.31\linewidth]{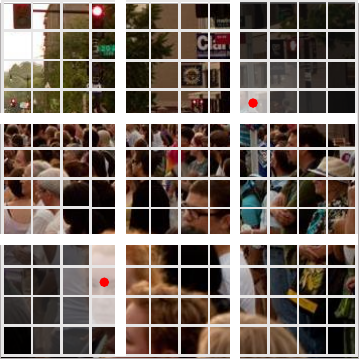}&
    \includegraphics[width=0.31\linewidth]{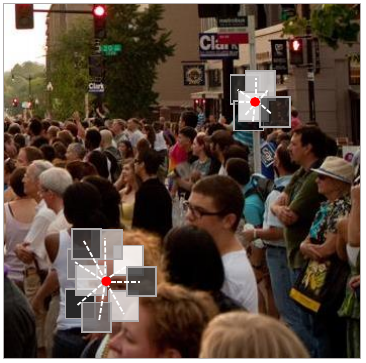}\\
    \rowfont{\scriptsize}
    (d) Global Attention & (e) Windowed Attention & (f) Deformable Attention\\
  \end{tabular}
  \caption{\textbf{Comparison of different feature representation schemes.} The sub-figure (a), (b) and (c) shows three convolutional schemes with regular, dilated and deformable kernels, respectively. Their weights are static to all spatial locations. (d) The global attention aggregates over all locations which is computational expensive. (e) Windowed attention works in a fixed-size window which is not adaptive to the layout of scene.
  (f) We introduce deformable attention for crowd counting for adaptive feature representation.}
  \label{fig:scale}
\end{figure}

Counting people in the crowd is challenging due to severe occlusions, various scene layouts, large scale variations of objects, \etc.
Over the past few years, a number of crowd counting methods~\cite{li2018csrnet,ma2019bayesian,wang2020distribution,liu2019adcrowdnet} have been proposed in the literature.
State-of-the-art approaches usually adopt convolutional neural networks (CNNs) to predict a density map for an input image. Then the total count is obtained by accumulating all the density values.
Although significant progress has been made in this research field, the traditional CNNs have inherent limitations due to their regular and limited receptive fields.
Concretely, a fixed-size (and fixed-weight) convolutional kernel is applied to all the spatial locations, regardless of geometric and visual content variations of crowds, as shown in~\Fig{fig:scale}~(a).
To enlarge the receptive fields, CSRNet~\cite{li2018csrnet} uses dilated convolutional kernels as illustrated in~\Fig{fig:scale}~(b).
However, the kernel size and weight are still fixed to the scene.
In order to adapt to geometric variations, deformable convolution~\cite{dai2017deformable} is introduced to crowd counting~\cite{liu2019adcrowdnet,guo2019dadnet}.
Deformable convolution adds 2D offsets to the regular grid sampling locations in the standard convolution.
It enables free-form deformation of the sampling grid to adaptively determinate the size of receptive field, as shown in~\Fig{fig:scale}~(c).
In general, effective deep architectures with adaptabilities to various scenes are essential to crowd counting.

Recently, Transformer architectures have received tremendous interest in image and vision tasks.
A number of methods based on Transformers have made impressive performances in image classification~\cite{dosovitskiy2021an,touvron21a,Yuan_2021_ICCV}, object detection~\cite{carion2020end}, image segmentation~\cite{Zheng_2021_CVPR}, \etc.
Compared to the widely-used CNNs in visual perception, Transformers enjoy great flexibility from the attention mechanism to dynamically model long-range dependencies in the image as shown in~\Fig{fig:scale}~(d).
However, a major problem of Transformers in vision tasks is the heavy computational complexity incurred by the spatial self-attention operation, which is quadratic to image size.
To reduce computation complexity, several methods~\cite{Liu_2021_ICCV,chu2021twins} are proposed to compute self-attention within local windows that partition an image, \ie, the windowed attention illustrated in~\Fig{fig:scale}~(e).
On the other hand, this scheme sacrifices the flexibility to handle variations in the scale of visual entities, which is especially disadvantageous to the counting task due to the large scale variations in crowds.

To address such limitations of fixed-size attention, this paper proposes a flexible Transformer architecture for crowd counting.
Specifically, we introduce \emph{Deformer}, a deformable attention in-built Transformer backbone to adaptively handle scale and layout variations in the scene.
There are two kinds of attention operators in the backbone, deformable attention~\cite{zhu2020deformable} and global attention.
For each spatial location, deformable attention aggregates surrounding features with learnable sampling offsets and attention weights as illustrated in~\Fig{fig:scale}~(f).
Due to this dynamic nature, deformable attention is helpful to fit the perspective distortion of the crowds, in contrast to windowed attention.
Then the global attention is engaged to build interactions between each feature element.
Deformer alternatively performs deformable and global attention, which facilitates representation learning adaptively to the image content.
We show that the features learned by Deformer are adaptive to the scale of crowds (see Fig.~\ref{fig:samplingpoints}).
Besides, it also enjoys a good trade-off between model capacity and computation efficiency due to the alternation of the two attention mechanisms.

In addition to the deformer backbone, we propose two modules to improve feature representation and density estimation further.
First, we propose the \emph{multi-level feature fusion} (MFF) module to fuse multi-scale features of Deformer.
Specifically, MFF adopts the deformable attention to efficiently fuse low-level fine-grained features with high-level semantic features,
leading to better counting and localization performance.
Furthermore, we propose the \emph{count-attentive feature enhancement} (CAFE) module to re-calibrate features under the global image context.
Given a count query, CAFE uses a transformer decoder to perceive the overall crowdedness of the scene.
Meanwhile, the intermediate multi-head attention maps are engaged to re-calibrate feature maps produced by the MFF module.
In this way, we enhance the foreground features while attenuating attention to the backgrounds (see Fig.~\ref{fig:attention}).
Finally, a lightweight prediction head is employed to estimate the density map.

Our main contributions are summarized as follows:

\begin{itemize}

\item
We propose a novel scene-adaptive attention network, \ie, SAANet, for crowd counting.
Our method enables learning both deformable sampling locations and dynamical attention weights, which is advantageous to convolution networks and Transformers with windowed attention in terms of flexibility for scale and scene variations.

\item
We design the multi-level feature fusion and count-attentive feature enhancement modules to strengthen feature representation further.
The learned representations are attentive to the foreground and adaptive to different scales of crowds, further facilitating the density map estimation.

\item
Extensive experiments demonstrate that our method achieves state-of-the-art performance on widely-used benchmarks, including NWPU-Crowd, UCF-QNRF, ShanghaiTech Part A\&B, and JHU-Crowd++.
On NWPU-Crowd, our method substantially reduces MAE compared with CNN-based methods and ranks No.1 on the public leaderboard.

\end{itemize}

%% file: sec/2_related_work.tex
\section{Related work}
\label{sec:related}

\subsection{Crowd Counting}
Crowd counting aims to estimate the number of people within a scene.
The early works in crowd counting are based on pedestrian detection~\cite{lin2001estimation,viola2005detecting, leibe2005pedestrian,wu2005detection,li2008estimating}. 
Due to crowded occlusion, the detection methods encounter obstacles and require fine-grained bounding box annotations. 
Direct regression methods~\cite{liu2015bayesian,wang2015deep,shang2016end,chattopadhyay2017counting} are then proposed to learn the global count directly. 
However, these methods do not incorporate spatial annotations, and the performances are unsatisfactory.
In contrast, recent crowd counting method~\cite{ma2019bayesian,wang2020distribution,ma2021learning,Wan_2021_CVPR} are mainly based on density regression, where the model firstly produces a density map of the scene and then provides the count estimation by summing the density values.
Thanks to the successful application and remarkable progress of convolution neural networks (CNNs), most of these methods in this field are based on the CNN architecture. 
Among them, various techniques and architectures are developed to improve the performance, including multi-scale architectures~\cite{Zhang_2016_CVPR,li2018csrnet,sindagi2017generating,yang2020embedding,song2021choose}, attention mechanism~\cite{mnih2014recurrent,bahdanau2014neural,vaswani2017attention}, loss function design~\cite{ma2019bayesian,wang2020distribution,ma2021learning,Wan_2021_CVPR}, \etc.

\subsubsection{CNN architectures}
Various architectures are proposed to cope with large scale variations of crowds.
MCNN~\cite{Zhang_2016_CVPR} propose a multi-column CNN where each column CNN utilizes filters with different receptive fields to adapt to scale variations.
CSRNet~\cite{li2018csrnet} uses dilated kernels to deliver larger reception fields and aggregate the multi-scale contextual information in the congested scenes. 
CP-CNN~\cite{sindagi2017generating} explicitly incorporates global and local contextual information for generating high-quality crowd density maps. 
Liu~\etal~\cite{Liu_2019_CVPR} propose a context-aware network to encode the contextual information, which are necessary for predicting accurate crowd densities.
Techniques like dilated convolution~\cite{guo2019dadnet,bai2020adaptive,yan2021crowd} and deformable convolution~\cite{zou2018net,guo2019dadnet,Liu_2019_CVPR} are also adopted to obtain larger or more flexible receptive fields.

\subsubsection{Attention mechanism}
Another useful technique is to incorporate attention mechanism~\cite{Liu_2019_CVPR,rong2021coarse,pan2020attention} into architectures. 
Liu~\etal~\cite{Liu_2019_CVPR} propose an attention-injective deformable convolutional network that designs a visual attention mechanism for alleviating the effects from various noises in congested scenes.
Rong~\etal~\cite{rong2021coarse} 
devise a coarse-to-fine attention mechanism by fusing multi-level features to better focus on the crowd area.

\subsubsection{Loss function and others}
More recent works focus on loss function design. Ma~\etal~\cite{ma2019bayesian} constructs a probability model and proposes Bayesian loss to learn density maps from annotation points directly. Wang~\etal~\cite{wang2020distribution} formulates crowd counting as a distribution matching problem and constructs loss using optimal transport. Lin~\etal~\cite{lin2021direct} further improve the loss function based on Sinkhorn distance. 
More improvements such as incorporating perspective information~\cite{yan2021crowd,yang2020reverse}, auxiliary task~\cite{wang2021self,yang2020embedding}, cross-datasets training~\cite{ma2021towards,chen2021variational} and neural architecture search~\cite{hu2020count} further promote the counting performance. 
However, as revealed in ~\cite{sun2021boosting,tian2021cctrans}, designing powerful deep architectures remains an active topic in crowd counting.

\subsection{Vision Transformer}
Recently, Transformer networks~\cite{vaswani2017attention} have become the dominant model for various natural language processing (NLP) tasks due to their outstanding performance in modeling long-range dependencies. Motivated by the success in NLP tasks, many attempts have been made to adapt transformer networks in vision tasks. The pioneering work of ViT~\cite{dosovitskiy2021an} directly applied a transformer on non-overlapping image patches for image classification and achieved remarkable performance. Since then, many works migrate transformer architectures into a variety of complex and challenging vision tasks such as image classification~\cite{dosovitskiy2021an,Liu_2021_ICCV,wang2021pyramid,chu2021twins}, object detection~\cite{carion2020end,zhu2020deformable}, semantic segmentation~\cite{Zheng_2021_CVPR} and image retrieval~\cite{elnouby2021training}.
There are few works~\cite{ranjan2019crowd,sun2021boosting,tian2021cctrans} attempt to introduce transformer-like architectures into crowd counting. Ranjan~\etal~\cite{ranjan2019crowd} employed a kind of self-attention mechanism for extracting non-local features and then combining with local CNN features to estimate the crowd density map. Concurrent to our work,
Sun~\etal~\cite{sun2021boosting} and Tian~\etal~\cite{tian2021cctrans} adopt the off-the-shelf T2T-ViT~\cite{Yuan_2021_ICCV} and Twins~\cite{chu2021twins} networks as backbone to extract features for crowd counting. These work highlights the bright prospects of introducing transformers into crowd counting.

%% file: sec/3_method.tex
\section{Methodology}
\label{sec:method}

\begin{figure*}
  \centering
  \includegraphics[width=\linewidth]{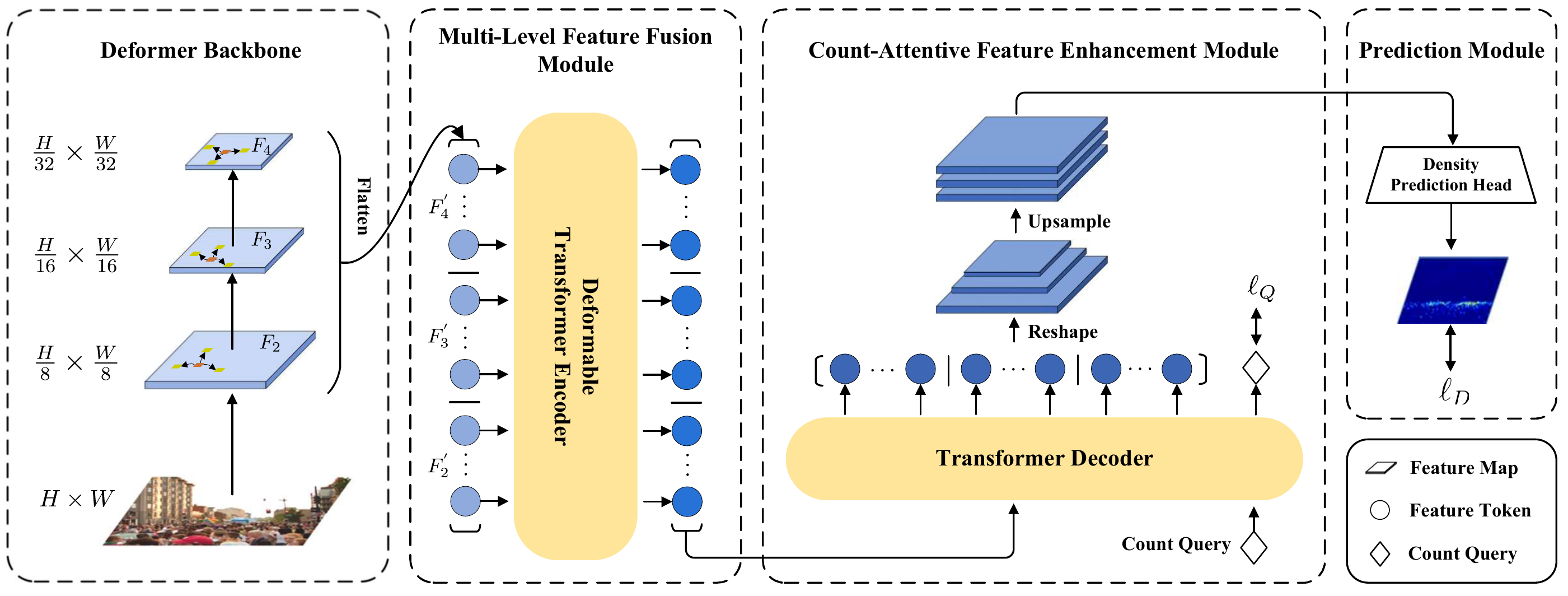}
  \caption{\textbf{The overview of SAANet.} The Deformer backbone first extracts multi-level features from given crowd image. Then, the multi-level feature fusion module aggregates these features by subsequently feeding them into a deformable transformer encoder. Then a count-attentive feature enhancement module re-calibrates the multi-level feature maps with the global context. Finally, the features are mapped to density map by a convolutional regression head.}
  \label{fig:dcc}
\end{figure*}

The basic idea of our method is to improve the flexibility and capacity to deal with scale and scene variations in crowd counting.
For this purpose, we design a framework named Scene-Adaptive Attention Network (SAANet) shown in~\Fig{fig:dcc}. 
In this section, we first introduce the overall architecture and then present the details of each module respectively.

\subsection{Overall Architecture}
The overall architecture of the proposed SAANet is depicted in~\Fig{fig:dcc}.
It consists of four main components: (1) a flexible backbone (\ie, Deformer) network, (2) a multi-level feature fusion (MFF) module, (3) a count-attentive feature enhancement (CAFE) module, and (4) a density map prediction head.

Given an image, we first input it to the Deformer backbone to extract multi-level features with resolutions of $\left\{ \frac{1}{8}, \frac{1}{16}, \frac{1}{32} \right\}$ of the original image.
We then feed the feature maps to the multi-level feature fusion module to fuse information across different levels.
Next, we employ the count-attentive feature enhancement module to leverage the global context information for feature enhancement. Finally, a prediction is engaged to regress the density map.
The details of each component are described in the following subsections.

\begin{table*}[ht]
  \centering
  \caption{\textbf{Detailed settings of Deformer series.} $P_i$ : patch size of Patch Embedding/Merging in the i-th Stage. $C_i$: channel dimension of feature maps in the i-th Stage. DA: deformable attention module. GSA: global self-attention module.}
  \label{tab:backbone}
  \def\arraystretch{1.0}
  \setlength{\tabcolsep}{12pt}
  \begin{tabular}{l|c|c|c|c|c}
    \toprule
    \multicolumn{1}{c|}{\multirow{2}{*}{Stage}} & \multirow{2}{*}{Output Size} & \multirow{2}{*}{Layer Name} & \multicolumn{3}{c}{Deformer Series} \\
    \cmidrule{4-6}
    & & & Deformer-tiny & Deformer-small & Deformer-base \\
    \midrule

    \multirow{3}{*}{Stage 1} & \multirow{3}{*}{$\frac{H}{4} \times \frac{W}{4}$} & Patch Embedding & $P_1 = 7$, $C_1 =96$ & $P_1 = 7$, $C_1 =96$ & $P_1 = 7$, $C_1 =128$ \\
    \cmidrule{3-6}
    & & \multirow{2}{*}{Encoder Block 1 w/ PE} & \multirow{2}{*}{$\begin{bmatrix} DA \end{bmatrix} \times 2$} & \multirow{2}{*}{$\begin{bmatrix} DA \end{bmatrix} \times 2$} & \multirow{2}{*}{$\begin{bmatrix} DA \end{bmatrix} \times 2$} \\
    & & & & &\\
    \midrule
    \multirow{3}{*}{Stage 2} & \multirow{3}{*}{$\frac{H}{8} \times \frac{W}{8}$} & Patch Merging & $P_2 = 3$, $C_2 = 192$ & $P_2 = 3$, $C_2 = 192$ & $P_2 = 3$, $C_2 = 256$ \\
    \cmidrule{3-6}
    & & \multirow{2}{*}{Encoder Block 2 w/ PE} & \multirow{2}{*}{$\begin{bmatrix} DA \end{bmatrix} \times 2$} & \multirow{2}{*}{$\begin{bmatrix} DA \end{bmatrix}\times2$} & \multirow{2}{*}{$\begin{bmatrix} DA \end{bmatrix} \times 2$} \\
    & & & & &\\
    \midrule
    \multirow{3}{*}{Stage 3} & \multirow{3}{*}{$\frac{H}{16} \times \frac{W}{16}$} & Patch Merging & $P_3 = 3$, $C_3 = 384$ & $P_3 = 3$, $C_3 = 384$ & $P_3 = 3$, $C_3 = 512$ \\
    \cmidrule{3-6}
    & & Encoder Block 3 w/ PE & $ \begin{bmatrix} DA \\GSA \end{bmatrix} \times 3$ & $ \begin{bmatrix} DA \\GSA \end{bmatrix} \times 9 $ & $ \begin{bmatrix} DA \\GSA \end{bmatrix} \times 9$ \\
    \midrule
    \multirow{3}{*}{Stage 4} & \multirow{3}{*}{$\frac{H}{32} \times \frac{W}{32}$} & Patch Merging & $P_4 = 3$, $C_4 = 768$ & $P_4 = 3$, $C_4 = 768$ & $P_4 = 3$, $C_4 = 1024$ \\
    \cmidrule{3-6}
    & & Encoder Block 4 w/ PE & $\begin{bmatrix} DA \\GSA \end{bmatrix} \times 1 $ & $\begin{bmatrix} DA \\GSA \end{bmatrix} \times 1 $ & $\begin{bmatrix} DA \\GSA \end{bmatrix} \times 1$ \\
    \bottomrule
  \end{tabular}
\end{table*}

\subsection{Deformer Backbone}
We aim to design a flexible transformer backbone to solve the scale and scene variations in crowd counting.
Different from the original ViT~\cite{dosovitskiy2021an} that can only generate a single resolution feature map, our backbone, termed Deformer, follows a hierarchical network structure that includes four stages.
The resolution is gradually reduced from stage 1 to stage 4 through patch embedding/merging (Section~\ref{sec:patch-mergeing}).
Each stage contains several encoder blocks, and there are two kinds of blocks: the global attention (Section~\ref{sec:global-attention}) and deformable attention (Section~\ref{sec:deformable-attention}) blocks.
In particular, we only use the deformable attention in the first two stages and perform these two attentions alternately in the last two stages.
There are two considerations for this design.

\begin{itemize}
  \item The first two stages contain more low-level features with higher resolutions. Since deformable attention is a kind of local attention, the calculation complexity is much smaller than global attention, especially in high-resolution features.
  
  \item In the last two stages, global attention can efficiently enlarge the receptive field to absorb global image context. Meanwhile, the computation complexity is acceptable since the spatial resolution is substantially reduced.
\end{itemize}

\subsubsection{Patch embedding/merging}
\label{sec:patch-mergeing}
Given an image with a resolution of $H \times W \times 3$, we perform patch embedding/merging to get a hierarchical feature map $F_{i}$ with a resolution of $\frac{H}{2^{i+1}}\times\frac{W}{2^{i+1}}\times C_i$, where $i\in\{1,2,3,4\}$.
We use a convolution layer (with kernel size $P$ and stride $S$) to perform patch merging in each stage.
To better maintain local continuity, we use overlapped patch merging.
Specifically, the first stage adopts a $P=7$ kernel with stride $S=4$ for patch embedding.
For other three stages, the kernel size is $P=3$ and stride is $S=2$ for patch merging.

\subsubsection{Global attention}
\label{sec:global-attention}

In global attention, the attention weights are calculated among all locations within a feature map, which is computationally expensive. Given an input feature map $\boldsymbol{x} \in \mathbb{R}^{N \times d}$, let $q \in \Omega_{q}$ indexes a query element with representation feature $\boldsymbol{z}_{q} \in \mathbb{R}^{d}$, and $k \in \Omega_{k}$ indexes a key element with representation feature $\boldsymbol{x}_k \in \mathbb{R}^{d}$, where $N$ means the spatial size of the feature map and $d$ is the feature dimension. $\Omega_{q}$ and $\Omega_{k}$ denote the set of query and key elements, respectively. Then the global self-attention is calculated as:
\begin{equation}
  \label{eq:globalattn}
  \text{GlobalAttn}(\boldsymbol{z}_q, \boldsymbol{x}) = \sum_{k \in \Omega_{k}} {A_{qk} \cdot \boldsymbol{x}_k^{T} W_V},
\end{equation}
where $W_V \in \mathbb{R}^{d \times d}$ is learnable weights. The attention weights $A_{qk} \propto \text{exp} \{ \boldsymbol{z}_q^T W_Q (\boldsymbol{x}_k^T W_K)^T / \sqrt{d} \}$
are normalized as $\sum_{k \in \Omega_{k}} A_{qk} = 1$, in which $W_Q$, $W_K \in \mathbb{R}^{d \times d}$ are also learnable weights.
Equ.~\eqref{eq:globalattn} can be extended to multi-head attention where each head attends to information from a subspace of the representation~\cite{vaswani2017attention}.

To incorporate the spatial information of scene layout, we perform positional encoding in each attention layer.
In particular, we add the positional
information into the input token before feeding into the attention layer, while the information is encoded by a $3 \times 3$ depth-wise convolution~\cite{chollet2017xception} with stride being 1.

\begin{figure}
  \centering
  \includegraphics[width=\linewidth]{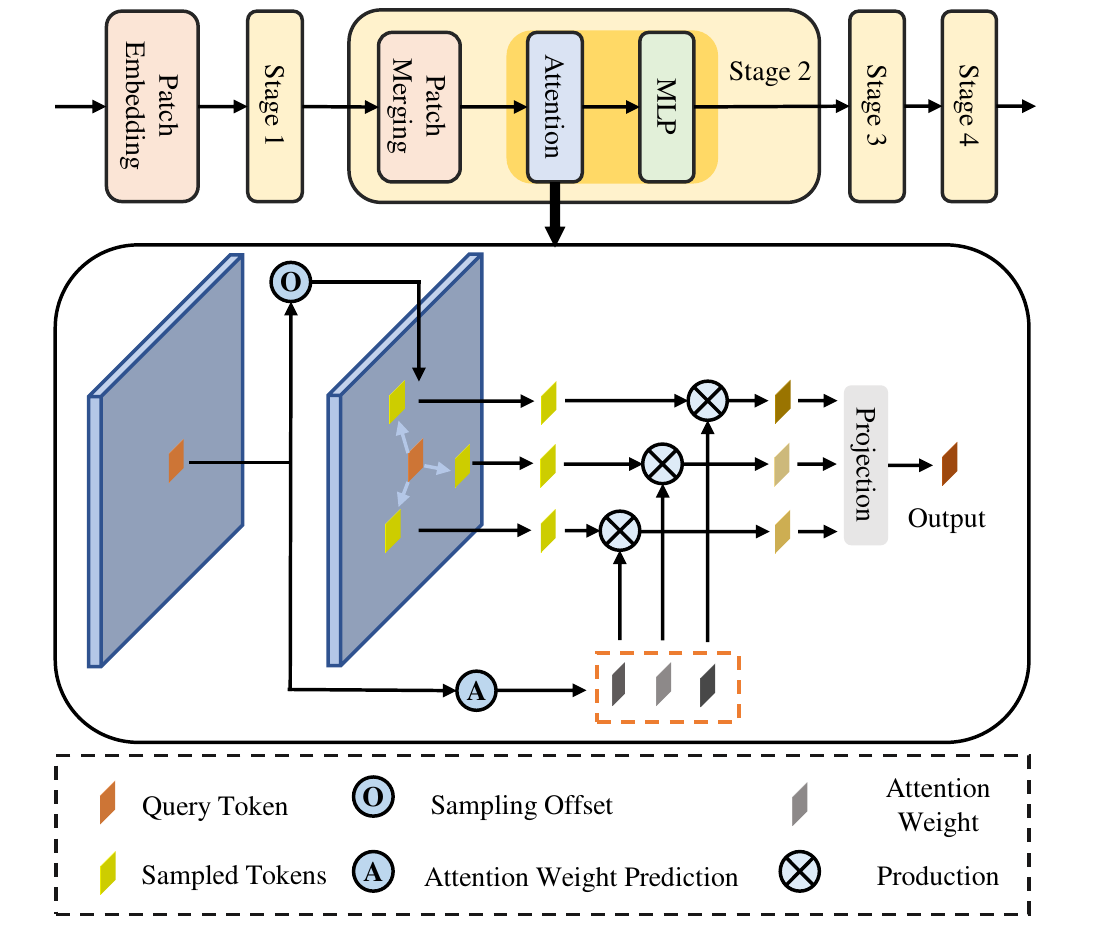}
  \caption{\textbf{Illustration of the architecture of Deformer and deformable attention.} The Deformer backbone contains four stages and patch embedding/merging is applied at the beginning of each stage. The deformable attention predicts both sampling locations and attention weights for a given query, and then use bilinear interpolation to get the sampled keys.}
  \label{fig:de_attn}
\end{figure}

\subsubsection{Deformable attention}
\label{sec:deformable-attention}

Instead of calculating the attention score globally, deformable attention only looks over a subset of keys decided by a query token, and the attention weight is dynamically generated.
In comparison with the windowed attention, it is also flexible on the scale of crowds.
We depict deformable attention in Fig~\ref{fig:de_attn}.
Formally, given an input feature map $\boldsymbol{x} \in \mathbb{R}^{N \times d}$, let $q$ index a query element with content feature $\boldsymbol{z}_q$ and a $2d$ reference point $p_q$, the deformable attention is calculated by
\begin{equation}
  \label{eq:deformattn}
  \text{DeformAttn}(\boldsymbol{z}_q, p_q, \boldsymbol{x}) = \sum_{k=1}^{K} {A_{qk} \cdot \boldsymbol{x}(p_q + \Delta p_{qk})^T W_V},
\end{equation}
where $k$ indexes the sampling point, and $K$ is the total sampled key number.
$\Delta p_{qk}$ and $A_{qk}$ denote the sampling offset and attention weight of the $k^{th}$ sampling point, which are both obtained by linear projection over the query feature $\boldsymbol{z}_q$.
The scalar attention weight $A_{qk}$ is normalized by $\sum_{k=1}^{K} A_{qk} =1$.
As the $p_q + \Delta p_{qk}$ locates the position of sampling point and is fractional, bilinear interpolation is applied to compute $\boldsymbol{x}(p_q + \Delta p_{qk})$.
Since $K \ll N$, the complexity of deformable attention is significantly lower than the global attention.
In this way, the attention mechanism generates dynamic weights to adapt various scenes, which is a key difference to deformable convolution~\cite{dai2017deformable}.

\subsubsection{Variants of Deformer}
We design a series of Deformers, Deformer-tiny to Deformer-base, with the same architecture but different sizes.
The overall structures of the Deformer series are summarized in~\Table{backbone}.
Deformer-tiny is a lightweight model for efficient inference, while Deformer-base is our best performed model with comparable complexity to other state-of-the-art networks in terms of FLOPs and parameters (see Table~\ref{tab:backbone_results} and Sec.~\ref{sec:ablation-deformer-backbone} for a detailed comparison).

\subsection{Multi-Level Feature Fusion}

Multi-level feature maps are important for the crowd counting task \cite{idrees2013multi,Zhang_2016_CVPR,yang2020embedding}.
Features in higher-level contain more semantics than those in lower-level, while the latter have more textural information and higher resolution, which are especially useful for congested areas.
As illustrated in Fig.~\ref{fig:dcc}, we use the output feature maps from the last three stages of the backbone for feature fusion, \ie, $F_{2} \in \mathbb{R}^{ \frac{H}{8} \times \frac{W}{8} \times C_{2}}$, $F_{3}\in \mathbb{R}^{ \frac{H}{16} \times \frac{W}{16} \times C_{3}}$ and $ F_{4}\in \mathbb{R}^{ \frac{H}{32} \times \frac{W}{32} \times C_{4}}$.
The multi-scale feature maps are projected to the ones with 256 channels by a linear projection layer and then flattened into feature tokens $F_{2}'\in \mathbb{R}^{ N_{2} \times 256}$, $F_{3}' \in \mathbb{R}^{ N_{3} \times 256}$, $F_{4}' \in \mathbb{R}^{ N_{4} \times 256}$, where the length of feature tokens $N_i = \frac{H}{2^{i+1}} \times \frac{W}{2^{i+1}}$.
Since the multi-head self-attention module has quadratic computation complexity to input size, we also employ the deformable attention module to implement the multi-level feature fusion.
We concatenate these feature tokens and input them into four deformable attention layers to fuse the features.
To identify which feature level each feature token lies in, we add a scale-level embedding in addition to the positional embedding.
In this way, low-level fine-grained features are fused with high-level semantic features.
Experimental results in the ablation study show that this module can effectively improve the counting accuracy.

\begin{figure}[t]
  \centering
  \includegraphics[width=\linewidth]{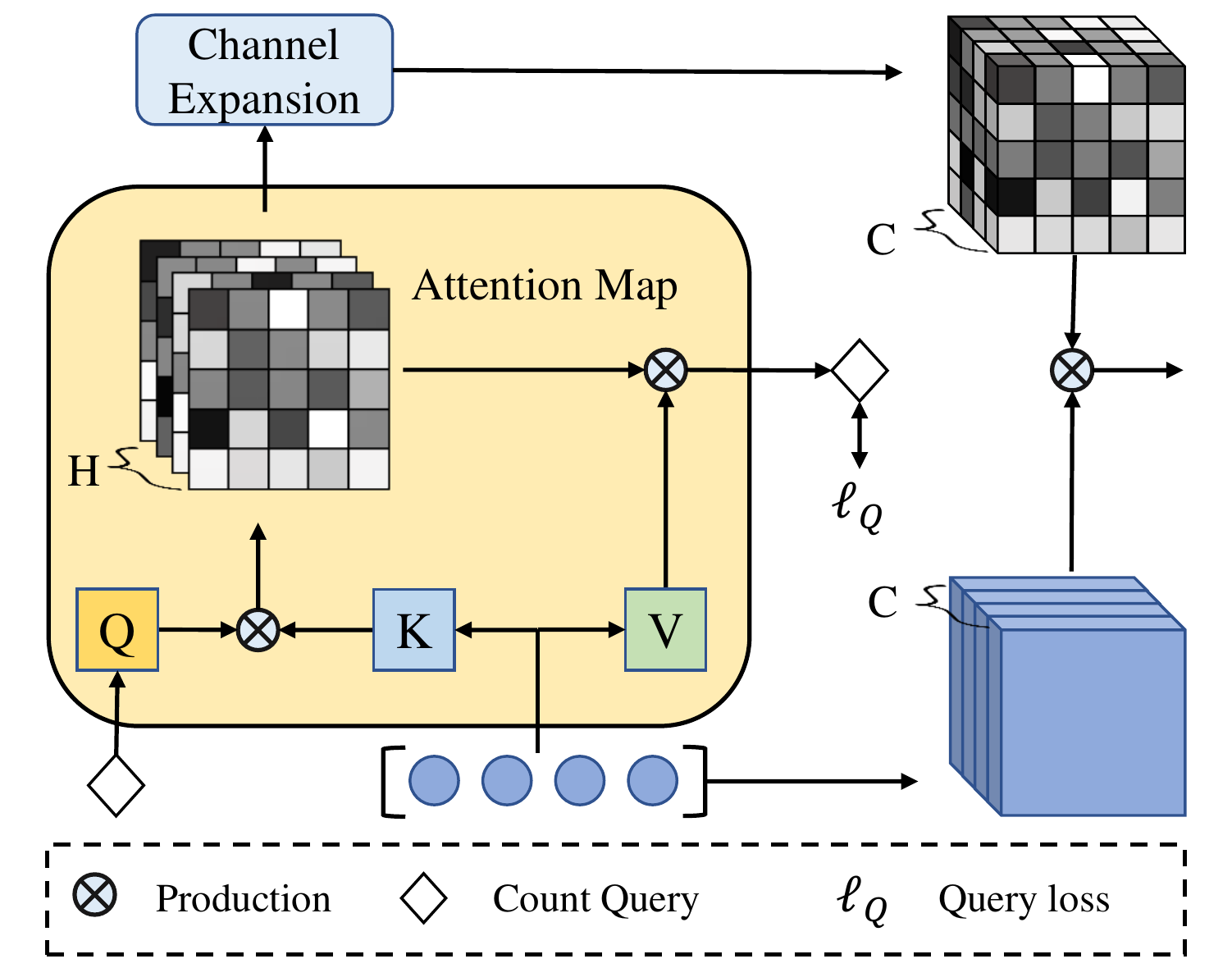}
  \caption{\textbf{Illustration of count-attentive feature enhancement.} Given the count query and the feature maps, a transformer decoder outputs the refined query features, which are used to regress the total count in the image. The multi-head attention maps produced in the cross attention layer are expanded in channel dimension to match the shape of the feature maps. Then we use the expanded attention maps to re-calibrate the feature maps via an element-wise production.}
  \label{fig:attnmap}
\end{figure}

\subsection{Count-Attentive Feature Enhancement}
After feature fusion, we design a transformer decoder to enhance feature representation further.
This module, named count-attentive feature enhancement (CAFE), employs a randomly initialized query to conduct cross attention with the multi-level feature maps.
The output query embedding could capture the global image context, which is used to regress the total count of the image.
Meanwhile, we use the multi-head attention maps in the cross-attention layer to re-calibrate feature maps.
Finally, a density map prediction head is adopted to estimate the density map.

\subsubsection{Query-based count regression}
We set a learnable count query $\boldsymbol{q}_{cnt}$ to incorporate the global information of the scene.
Specifically, as is illustrated in Fig.~\ref{fig:attnmap}, we input the count query and the refined multi-scale feature maps into a transformer decoder~\cite{vaswani2017attention} for cross-attention computation.
The number of decoder layers is set to 4 in the experiment.
Finally, we adopt the output query feature to predict the total count in the image, which serves as an auxiliary training term.

\subsubsection{Feature re-calibration}
The transformer decoder uses the count query to perceive the overall crowdedness of the scene.
Therefore, the multi-head attention maps produced by the cross attention layer can focus on the crowd region and ignore the backgrounds.
Naturally, we adopt the intermediate multi-head attention maps to re-calibrate the feature maps in both spatial and channel dimensions.
As shown in~\Fig{fig:attnmap}, we first expand the channels of the attention maps to align with the feature maps.
Then we perform element-wise production to produce the re-calibrated feature maps.
In this way, we enhance the foreground features while attenuating attention to the backgrounds.

\subsection{Density Prediction Head and Loss Function}
\subsubsection{Density prediction head}
We adopt a simple regression head to regress accurate density maps. 
Specifically, the density regression head consists of two $3\times3$ convolution layers and one $1\times1$ convolution layer. The first two layers aim to shrink the channels gradually, while the last layer regresses a density map directly.

\subsubsection{Loss function}
The overall loss consists of two parts. First, we adopt distribution matching loss $\ell_{D}$ proposed in \cite{wang2020distribution} to supervise the density map regression and count estimation.
Additionally, we use an auxiliary $L_1$ loss term (namely query loss $\ell_Q$) to supervise the count regression in the aforementioned count-attentive feature enhancement module.
The overall loss function can be formulated as follows:
\begin{equation}\label{equ:loss}
    \ell = \ell_{D}(D, D^{*}) + \ell_Q(C, C^{*}),
\end{equation}
where $D$ and $D^{*}$ represent the predicted density map and annotation map, respectively, $C$ denotes the count prediction in the CAFE module, and $C^{*}$ represents the ground truth count.
At inference, the predicted count is given by summation over the predicted density map, \emph{i.e.}, $\Vert D \Vert_1$.

%% file: sec/4_experiments.tex
\section{Experiments}
\label{sec:experiment}

In this section, we conduct extensive experiments and ablation studies on crowd counting benchmarks to verify the effectiveness of our proposed method and compare with state of the arts.

\begin{table*}[t]
	\centering
	\caption{Performance comparison on SHA, SHB and UCF-QNRF datasets.}
	\label{tab:counting_results}
	\setlength{\tabcolsep}{12pt}
	\def\arraystretch{1.1}
	\begin{tabular}{r|c|c|cc|cc|cc}
		\toprule
		\multirow{2}{*}{Method} & \multirow{2}{*}{Venue} & \multirow{2}{*}{Backbone} & \multicolumn{2}{c|}{SHA} & \multicolumn{2}{c|}{SHB} & \multicolumn{2}{c}{UCF-QNRF} \\
		\cmidrule{4-9} & & & MAE & MSE & MAE & MSE & MAE & MSE \\
		\midrule
		MCNN~\cite{Zhang_2016_CVPR}      & CVPR16  & Custom CNN & 110.2 & 173.2 & 26.4 & 41.3 & 277.0 & 426.0 \\
		CSRNet~\cite{li2018csrnet}       & CVPR18  & VGG-16     & 68.2 & 115.0 & 10.6 & 16.0 & - & - \\
		CAN~\cite{liu2019context}        & CVPR19  & VGG-16     & 62.3 & 100.0 & 7.8  & 12.2 & 107.0 & 183.0 \\
		SFCN~\cite{wang2019learning}     & CVPR19  & ResNet-101 & 64.8 & 107.5 & 7.6  & 13.0 & 102.0 & 171.4 \\
		PACNN~\cite{shi2019revisiting}   & CVPR19  & VGG-16     & 62.4 & 102.0 & 7.6  & 11.8 & - & - \\
		S-DCNet~\cite{xiong2019open}     & ICCV19  & VGG-16     & 58.3 & 95.0  & 6.7  & 10.7 & 104.4 & 176.1 \\
		DSSI-Net~\cite{liu2019crowd}     & ICCV19  & VGG-16     & 60.6 & 96.0  & 6.8  & 10.3 & 99.1 & 159.2 \\
		BL~\cite{ma2019bayesian}         & ICCV19  & VGG-19     & 62.8 & 101.8 & 7.7  & 12.7 & 88.7 & 154.8 \\
		CRNet~\cite{liu2020crowd}        & IEEE-TIP20 & ConvLSTM & 56.4 & 90.4 & 7.4 & 11.9 & 101.0 & 162.0 \\
		RPNet~\cite{yang2020reverse}     & CVPR20  & VGG-16     & 61.2 & 96.9  & 8.1  & 11.6 & - & - \\
		ASNet~\cite{jiang2020attention}  & CVPR20  & VGG-16     & 57.8 & 90.1  & -  &-   & 91.6 & 159.7 \\
		LibraNet~\cite{liu2020weighing}  & ECCV20  & VGG-16     & 55.9 & 97.1  & 7.3 & 11.3 & 88.1 & 143.7 \\
		AMRNet~\cite{liu2020adaptive}    & ECCV20  & VGG-16     & 61.5 & 98.3  & 7.0 & 11.0 & 86.6 & 152.2 \\
		DPN-IPSM~\cite{ma2020learning}   & MM20    & VGG-19     & 58.1 & 91.7  & 6.5 & 10.1 & 84.7 & 147.2 \\
		NoisyCC~\cite{wan2020modeling}   & NeurIPS20 & VGG-19   & 61.9 & 99.6 & 7.4 & 11.3 & 85.6 & 148.3 \\
		DM-Count~\cite{wang2020distribution} & NeurIPS20 & VGG-19 & 59.7 & 95.7 & 7.4 & 11.8 & 85.6 & 148.3 \\
		UOT~\cite{ma2021learning}        & AAAI21  & VGG-19     & 58.1 & 95.9 & 6.5 & 10.2 & 83.3 & 142.3 \\
		S3~\cite{lin2021direct}          & IJCAI21 & VGG-19     & 57.0 & 96.0 & 6.3 & 10.6 & 80.6 & 139.8 \\
		GL~\cite{Wan_2021_CVPR}          & CVPR21  & VGG-19     & 61.3 & 95.4 & 7.3 & 11.7 & 84.3 & 147.5 \\
		D2CNet~\cite{9347700}            & IEEE-TIP21 & VGG-16  & 57.2 & 93.0 & 6.3 & 10.7 & 81.7 & 137.9 \\
		P2PNet~\cite{song2021rethinking} & ICCV21  & VGG-16     & 52.7 & 85.1 & 6.3 & 9.9 & 85.3 & 154.5 \\
		SDA+DM~\cite{ma2021towards}      & ICCV21  & VGG-19     & 55.0 & 92.7 & - & - & 80.7 & 146.3 \\
		FDC~\cite{Liu_2021_ICCV}            & ICCV21  & ResNet-18  & 65.4 & 109.2 & 11.4 & 19.1 & 93.0 & 157.3\\
		MFDC~\cite{Liu_2021_ICCV}        & ICCV21  & ResNet-18  & 55.4 & 91.3 & 6.9 & 10.3 & \textbf{76.2} & \textbf{121.5}\\
		\midrule
		BCCT~\cite{sun2021boosting}      & arXiv21 & T2T-ViT-14 & 53.1 & 82.2 & 7.3 & 11.3 & 83.8 & 143.4 \\
		CCTrans~\cite{tian2021cctrans}   & arXiv21 & Twins-large & 52.3 & 84.9 & 6.2 & 9.9 & 82.8 & 142.3 \\
		SAANet (Ours)                    & -       & Deformer-base & \textbf{51.7}& \textbf{80.1} & \textbf{6.1} & \textbf{9.4} & 79.3 & 137.3 \\
		\bottomrule
	\end{tabular}
\end{table*}

\subsection{Datasets}
Four challenging crowd counting benchmarks are used to evaluate our method.

\subsubsection{ShanghaiTech}~\cite{Zhang_2016_CVPR} contains 1,198 images, with a total of 330,165 annotations.
The dataset is divided into two parts by the density of a single sample: Part A (henceforth, SHA) and Part B (SHB). SHA has a significantly higher density than SHB.
The images of SHA are randomly collected from the internet. SHA contains 482 images, and the number of annotations in an image ranges from 33 to 3139. The training set has 300 images, and the remaining 182 images form the testing set.
SHB contains 716 images taken in the crowded street of Shanghai.
All the images in SHB are with a fixed resolution, and the number of people in a single image ranges from 9 to 578. 
There are 316 images for training and 400 images in the test set.

\subsubsection{UCF-QNRF}~\cite{idrees2018composition} is a congested crowd dataset that contains 1535 high-resolution images crawled from Flickr, Web Search, and Hajj footage. It contains about 1.25 million point annotations.
The average number of persons per image is 815, and the maximum number reaches 12,865.
It is a challenging dataset since it has various scenes, perspectives, image resolutions, densities, and illumination conditions.
Among the dataset, 1201 images are used for training, and the remaining 334 images are used for testing. 

\subsubsection{NWPU-Crowd}~\cite{wang2020nwpu} is a large-scale and challenging datasets for crowd counting.
It contains 5109 images and 2.13 million annotations, of which 3109 images are used for training, 500 images are used for validation, and 1500 images are used for testing. The dataset has significant scale variations, where the number of annotated people in a single image ranges from $0$ to $20,033$ and contains diversified scenes. Moreover, the dataset introduces 351 negative samples of nobody or textures similar to congested crowd scenes, making it much more challenging. 
According to density level, the testing set is divided into the following fine-grained subsets, (1) S0: images of negative samples, (2) S1: images containing $1\sim100$ people, (3) S2: images with $101\sim500$ people, (4) S3: images with $501\sim 5000$ people and (5) S4: images contain more than 5000 people. According to the luminance condition (more specifically, images' average luminance values in the YUV color space), the dataset can be divided into three subsets (6) L0: luminance value between $[0, 0.25]$, (7) L1: luminance value between $(0.25, 0.5]$ and (8) L2: luminance value between $(0.5, 0.75]$. We study counting performance on all the subsets and the overall dataset.

\subsubsection{JHU-Crowd++}~\cite{sindagi2020jhu} is a large-scale dataset that contains 4372 images with 1.51 million annotations. There are 2272 images used for training, 500 images for validation, and the remaining 1600 images are used for testing.
The images are collected under diverse conditions and various geographical locations. 
Additionally, the dataset contains a number of images with weather degradations and illumination variations, which makes it a very challenging dataset. The testing dataset can be divided into several fine-grained sub-categories according to density level or conditions: (1) Low: images with $0\sim 50$ people, (2) Medium: images with $51\sim 500$ people, (3) High: images more than 500 people and 
(4) Weather: weather degraded images.
We present the counting performance on both the overall dataset and fine-grained sub-categories.

\begin{table*}[ht]
	\centering
	\caption{Performance comparison on NWPU-Crowd dataset. We present the counting performance on different density levels and luminance conditions.}
	\label{tab:nwpu_result}
	\def\arraystretch{1.1}
	\resizebox{\linewidth}{!}{
		\begin{tabular}{r|c|cc|ccc|c|ccccc|c|ccc}
			\toprule
			\multirow{3}{*}{Method} & \multirow{3}{*}{Venue} & \multicolumn{2}{c|}{Val} & \multicolumn{11}{c}{Test}
			\\\cmidrule{3-17}
			& & \multicolumn{2}{c|}{Overall} & \multicolumn{3}{c|}{Overall} & \multicolumn{6}{c|}{Scene Level (MAE)} & \multicolumn{4}{c}{Luminance (MAE)} \\
			\cmidrule{3-17}
			& & MAE & MSE & MAE & MSE & NAE & Avg. & S0 & S1 & S2 & S3 & S4 & Avg. & L0 & L1 & L2 \\
			\midrule
			MCNN~\cite{Zhang_2016_CVPR} & CVPR16 & \cellcolor{black!10}218.5 & \cellcolor{black!10}700.6 & \cellcolor{black!10}232.5 & \cellcolor{black!10}714.6 & \cellcolor{black!10}1.063 & 1171.9 & 356.0 & 72.1 & 103.5 & 509.5 & 4818.2 & 220.9 & 472.9 & 230.1 & 181.6 \\
			CSRNet~\cite{Li_2018_CVPR}  & CVPR18 & \cellcolor{black!10}104.9 & \cellcolor{black!10}433.5 & \cellcolor{black!10}121.3 & \cellcolor{black!10}387.8 & \cellcolor{black!10}0.604 & 522.7 & 176.0 & 35.8 & 59.8  & 285.8 & 2055.8 & 112.0 & 232.4 & 121.0 & 95.5 \\
			CAN~\cite{liu2019context}   & CVPR19 & \cellcolor{black!10}93.5  & \cellcolor{black!10}489.9 & \cellcolor{black!10}106.3 & \cellcolor{black!10}386.5 & \cellcolor{black!10}0.295 &612.2 &82.6  & 14.7 & 46.6  & 269.7 & 2647.0 & 102.1 & 222.1 & 104.9 & 82.3 \\
			BL~\cite{ma2019bayesian}    & ICCV19 & \cellcolor{black!10}93.6  & \cellcolor{black!10}470.4 & \cellcolor{black!10}105.4 & \cellcolor{black!10}454.2 & \cellcolor{black!10}0.203 & 750.5 & 66.5 & 8.7 & 41.2 & 249.9 & 3386.4 & 115.8 & 293.4 & 102.7 & 68.0 \\
			SFCN+~\cite{9153156}        & PAMI20 & \cellcolor{black!10}95.5  & \cellcolor{black!10}608.3 & \cellcolor{black!10}105.7 & \cellcolor{black!10}424.1 & \cellcolor{black!10}0.254 & 712.7 & 54.2 & 14.8 & 44.4 & 249.6 & 3200.5 & 106.8 & 245.9 & 103.4 & 78.8 \\
			DM-Count~\cite{wang2020distribution} & NeurIPS20 & \cellcolor{black!10}-& \cellcolor{black!10}-& \cellcolor{black!10}88.4 & \cellcolor{black!10}388.6 & \cellcolor{black!10}0.169 & 498.0 & 146.7 & 7.6 & 31.2 & 228.7 & 2075.8 & 88.0 & 203.6 & 88.1 & 61.2 \\
			UOT~\cite{ma2021learning}   & AAAI21 & \cellcolor{black!10}70.5& \cellcolor{black!10}357.6&  \cellcolor{black!10}87.8 & \cellcolor{black!10}387.5 & \cellcolor{black!10}0.185 & 566.5 & 80.7 & 7.9 & 36.3 & 212.0 & 2495.4 &95.2 &240.3 & 86.4 & 54.9 \\
			GL~\cite{Wan_2021_CVPR}     & CVPR21 & \cellcolor{black!10}- & \cellcolor{black!10}- & \cellcolor{black!10}79.3 & \cellcolor{black!10}346.1 & \cellcolor{black!10}0.180 & 508.5 & 92.4 & 8.2 & 35.4 & 179.2 & 2228.3 & 85.6 & 216.6 & 78.6 & 48.0 \\
			D2CNet~\cite{9347700}       & IEEE-TIP21& \cellcolor{black!10}- & \cellcolor{black!10}- & \cellcolor{black!10}85.5 & \cellcolor{black!10}361.5 & \cellcolor{black!10}0.221 & 539.9 & 52.4 & 10.8 &36.2 &212.2 &2387.8 & 82.0 & 177.0 & 83.9 & 68.2\\
			P2PNet~\cite{wu2021rethinking} & ICCV21 & \cellcolor{black!10}- & \cellcolor{black!10}- & \cellcolor{black!10}72.6  & \cellcolor{black!10}331.6 & \cellcolor{black!10}0.192 & 510.0 & 34.7 & 11.3 & 31.5 & 161.0 & 2311.6 & 80.6 & 203.8 & 69.6 & \textbf{50.1} \\
			FDC~\cite{Liu_2021_ICCV}   & ICCV21 & \cellcolor{black!10}- & \cellcolor{black!10}- & \cellcolor{black!10}119.4 & \cellcolor{black!10}380.6 & \cellcolor{black!10}0.340 & 642.7 & 81.4 & 20.4 & 59.6 & 295.5 & 2756.7 & 105.0 & 206.1 & 119.8 & 97.1\\
			MFDC~\cite{Liu_2021_ICCV}   & ICCV21 & \cellcolor{black!10}- & \cellcolor{black!10}- & \cellcolor{black!10}74.7 & \cellcolor{black!10}\textbf{267.9} & \cellcolor{black!10}0.184 & 412.2 & \textbf{5.7} & 8.5 & 33.3 & 216.0 & 1797.7 & 67.6 & 138.5 & 75.2 & 57.6\\
			\midrule
			BCCT~\cite{sun2021boosting} & arXiv21 & \cellcolor{black!10}53.0  & \cellcolor{black!10}170.3 & \cellcolor{black!10}82.0 & \cellcolor{black!10}366.9 & \cellcolor{black!10}0.164 & - & - & - & - & - & - & - & - & - & - \\
			CCTrans~\cite{tian2021cctrans} & arXiv21 & \cellcolor{black!10}38.6  & \cellcolor{black!10}87.8 & \cellcolor{black!10}69.3 & \cellcolor{black!10}299.4 & \cellcolor{black!10}0.135 & 429.5 & 59.0 & 6.3 &27.4 & 178.2 & 1876.6 & 65.5 & 137.6 & 67.7 & 57.5 \\
			SAANet (Ours)               & - & \cellcolor{black!10}\textbf{34.3} & \cellcolor{black!10}\textbf{79.8} &\cellcolor{black!10}\textbf{66.0}  & \cellcolor{black!10}298.8 & \cellcolor{black!10}\textbf{0.123} & \textbf{395.5}& 62.0& \textbf{5.3}& \textbf{26.6}& \textbf{171.9}& \textbf{1711.5}& \textbf{61.8}& \textbf{130.0}& \textbf{65.2}&53.1 \\
			\bottomrule
		\end{tabular}
	}
\end{table*}

\begin{table*}[ht]
	\centering
	\caption{Performance comparison on JHU-Crowd++ dataset. We present the counting performance on both overall dataset and fine-grained sub-categories.}
	\label{tab:jhu_result}
	\def\arraystretch{1.1}
	\begin{tabular}{r|c|cc|cc|cc|cc|cc|cc}
		\toprule
		\multirow{3}{*}{Method} & \multirow{3}{*}{Venue} & \multicolumn{2}{c|}{Val} & \multicolumn{10}{c}{Test}\\
		\cmidrule{3-14}
		& & \multicolumn{2}{c|}{Overall} & \multicolumn{2}{c|}{Overall} & \multicolumn{2}{c|}{Low} & \multicolumn{2}{c|}{Medium} & \multicolumn{2}{c|}{High} & \multicolumn{2}{c}{Weather} \\ 
		\cmidrule{3-14}
		& & MAE & MSE & MAE & MSE & MAE & MSE & MAE & MSE & MAE & MSE & MAE & MSE \\ 
		\midrule
		MCNN~\cite{Zhang_2016_CVPR}      & CVPR16  &\cellcolor{black!10}160.6 &\cellcolor{black!10}377.7 & \cellcolor{black!10}188.9 & \cellcolor{black!10}483.4 & 97.1 & 192.3 & 121.4 & 191.3 & 618.6 & 1,166.7 & 330.6 & 852.1  \\ 
		SANET~\cite{cao2018scale}        & ECCV18  & \cellcolor{black!10}82.1 &\cellcolor{black!10}272.6 & \cellcolor{black!10}91.1 & \cellcolor{black!10}320.4 & 17.3 & 37.9 & 46.8 & 69.1 & 397.9 & 817.7 & 154.2 & 685.7  \\ 
		CSRNET~\cite{Li_2018_CVPR}       & CVPR18  & \cellcolor{black!10}72.2 & \cellcolor{black!10}249.9 & \cellcolor{black!10}85.9 & \cellcolor{black!10}309.2 & 27.1 & 64.9 & 43.9 & 71.2 & 356.2 & 784.4 & 141.4 & 640.1 \\ 
		CAN~\cite{liu2019context}        & CVPR19  & \cellcolor{black!10}89.5& \cellcolor{black!10}239.3 & \cellcolor{black!10}100.1 & \cellcolor{black!10}314.0 & 37.6 & 78.8 & 56.4 & 86.2 & 384.2 & 789.0 & 155.4 & 617.0 \\ 
		SFCN~\cite{wang2019learning}     & CVPR19  & \cellcolor{black!10}62.9& \cellcolor{black!10}247.5 & \cellcolor{black!10}77.5 & \cellcolor{black!10}297.6 & 16.5 & 55.7 & 38.1 & 59.8 & 341.8 & 758.8 & 122.8 & 606.3 \\ 
		BL~\cite{ma2019bayesian}         & ICCV19  & \cellcolor{black!10}59.3 &\cellcolor{black!10}229.2 & \cellcolor{black!10}75.0 & \cellcolor{black!10}299.9 & 10.1 & 32.7 & 34.2 & 54.5 & 352.0 & 768.7 & 140.1 & 675.7 \\ 
		CG-DRCN-CC~\cite{sindagi2020jhu} & PAMI20  & \cellcolor{black!10}57.6 &\cellcolor{black!10}244.4 & \cellcolor{black!10}71.0 & \cellcolor{black!10}278.6 & 14.0 & 42.8 & 35.0 & 53.7 & 314.7 & 712.3 & 120.0 & 580.8 \\
		D2CNet~\cite{9347700}            & IEEE-TIP21   &\cellcolor{black!10}- &\cellcolor{black!10}- & \cellcolor{black!10}73.7 & \cellcolor{black!10}292.5 & 12.6 & 38.5 & 36.5 & 56.3 & 330.3 & 748.6 & 135.3 & 678.2 \\
		UOT~\cite{ma2021learning}        & AAAI21 & \cellcolor{black!10}- & \cellcolor{black!10}- & \cellcolor{black!10}60.5 & \cellcolor{black!10}252.7 & 11.2 & 26.2 & 28.7 & \textbf{45.3} & 274.1 & 648.2 & 114.9 & 610.7 \\
		SDA+DM~\cite{ma2021towards} & ICCV21 & \cellcolor{black!10}- & \cellcolor{black!10}- & \cellcolor{black!10}59.3 & \cellcolor{black!10}248.9 & - & - & - & - & - & - & - & - \\
		FDC~\cite{Liu_2021_ICCV} & ICCV21 & \cellcolor{black!10}- & \cellcolor{black!10}- & \cellcolor{black!10}77.8 & \cellcolor{black!10}263.1 & - & - & - & - & - & - & - & - \\
		MFDC~\cite{Liu_2021_ICCV} & ICCV21 & \cellcolor{black!10}- & \cellcolor{black!10}- & \cellcolor{black!10}58.1 & \cellcolor{black!10}221.9 & - & - & - & - & - & - & - & - \\
		\midrule
		BCCT~\cite{sun2021boosting}      & arXiv21 & \cellcolor{black!10}46.5 & \cellcolor{black!10}198.6 & \cellcolor{black!10}54.8 & \cellcolor{black!10}\textbf{208.5} & - & - & - & - & - & - & - & - \\ 
		SAANet (Ours)                    & -       & \cellcolor{black!10}\textbf{42.7} & \cellcolor{black!10}\textbf{186.4} & \cellcolor{black!10}\textbf{51.6}  & \cellcolor{black!10}214.0 & \textbf{6.0} & \textbf{16.3} & \textbf{27.2} & 45.6 & \textbf{227.6} & \textbf{547.0} & \textbf{93.9} & \textbf{506.4} \\ 
		\bottomrule
	\end{tabular}
\end{table*}

\subsection{Evaluation Criteria}
We adopt three popular metrics in crowd counting to evaluate our method: Mean Absolute Error (MAE), root Mean Squared Error (MSE) and mean Normalized Absolute Error (NAE).
These metrics can be formulated as follows:
\begin{equation}
MAE = \dfrac{1}{N}\sum_{i=1}^N|C_{i} - C_{i}^{gt}|,
\end{equation}
\begin{equation}
MSE = \sqrt{\dfrac{1}{N}\sum_{i=1}^N{|C_{i} - C_{i}^{gt}|}^2},
\end{equation}
\begin{equation}
NAE = \dfrac{1}{N}{\sum_{i=1}^N|C_{i} - C_{i}^{gt}|}\,/\,{C_{i}^{gt}},
\end{equation}
where $N$ is the number of testing images, $C_{i}$ and $C_{i}^{gt}$ are the estimated count and the ground truth count of the $i$-th image respectively. Since NWPU-Crowd has a number of negative samples with zero annotation, they are excluded during the calculation of NAE to avoid zero denominators.

\begin{figure*}[!t]
	\centering
	\setlength{\tabcolsep}{1pt}
	\begin{tabular}{ccccc}
		\small{GT: 1111} & \small{GT: 400} & \small{GT:236 } & \small{GT:  170} & \small{GT: 82}\\
		\includegraphics[width=0.195\linewidth,height=0.15\linewidth]{fig/fig5/IMG_2.jpg} & 
		\includegraphics[width=0.195\linewidth,height=0.15\linewidth]{fig/fig5/IMG_179.jpg} &
		\includegraphics[width=0.195\linewidth,height=0.15\linewidth]{fig/fig5/IMG_40.jpg} & 
		\includegraphics[width=0.195\linewidth,height=0.15\linewidth]{fig/fig5/IMG_4.jpg} &
		\includegraphics[width=0.195\linewidth,height=0.15\linewidth]{fig/fig5/IMG_5.jpg}\\
		\small{Pred: 1353.4} & \small{Pred: 358.5} & \small{Pred: 285.1} & \small{Pred: 156.7} & \small{Pred: 69.7}\\
		\includegraphics[width=0.195\linewidth,height=0.15\linewidth]{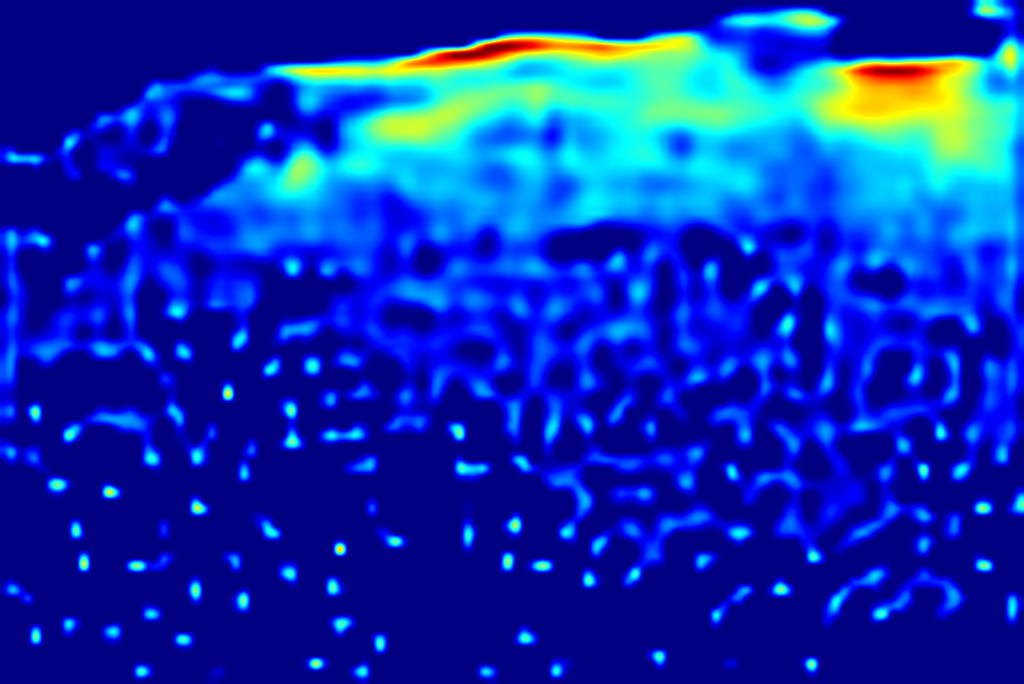} &
		\includegraphics[width=0.195\linewidth,height=0.15\linewidth]{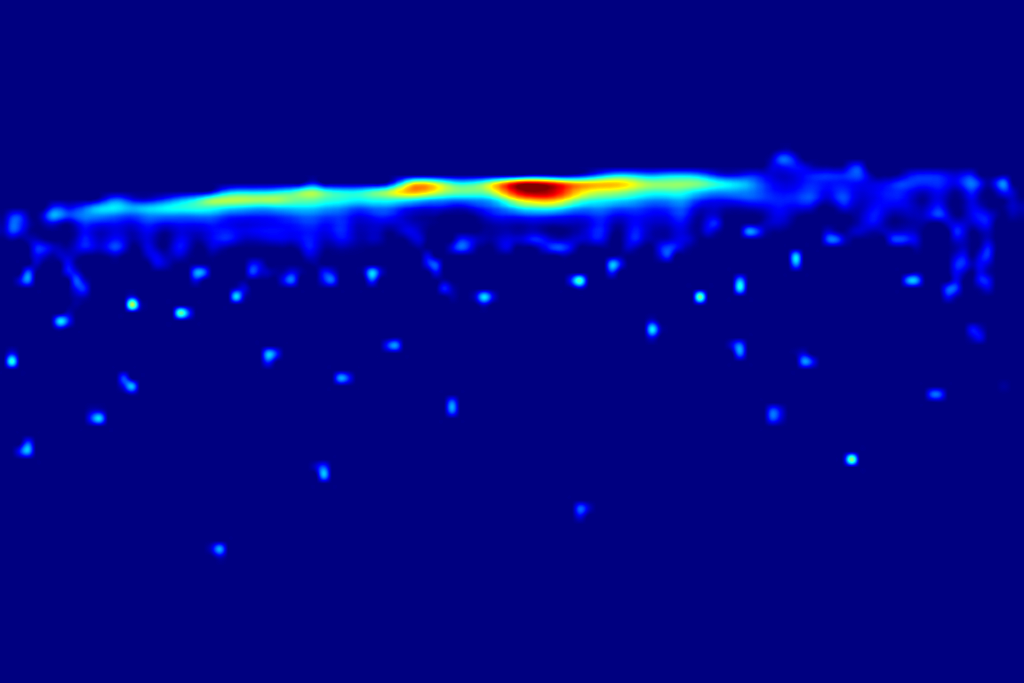} &
		\includegraphics[width=0.195\linewidth,height=0.15\linewidth]{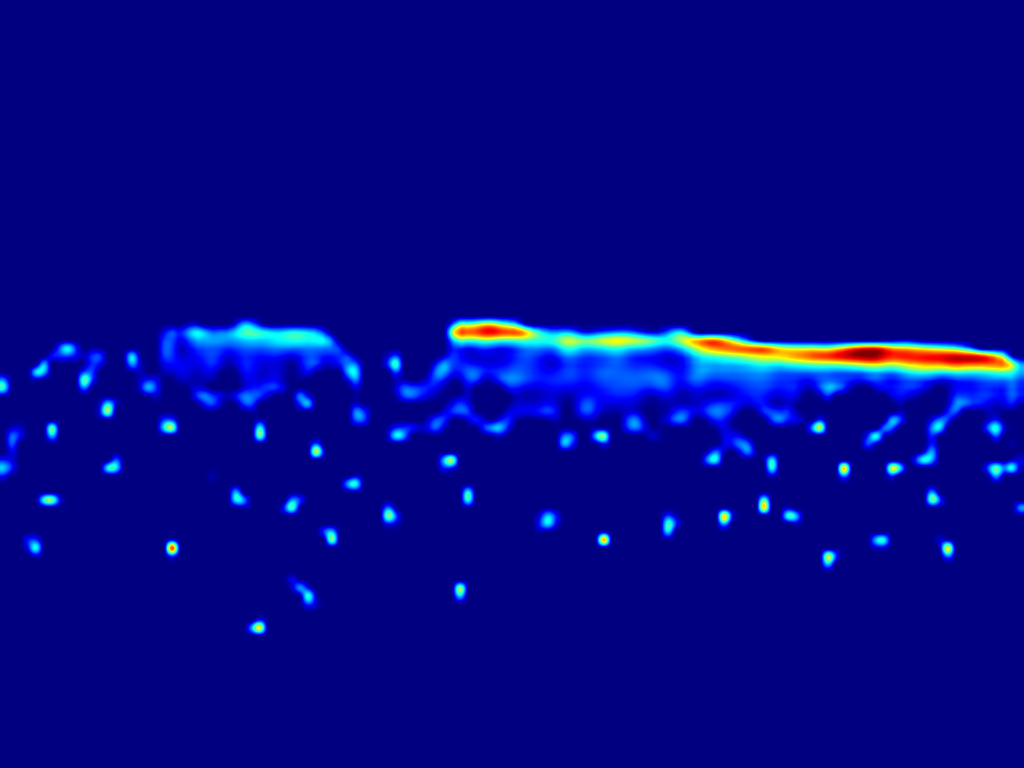} &
		\includegraphics[width=0.195\linewidth,height=0.15\linewidth]{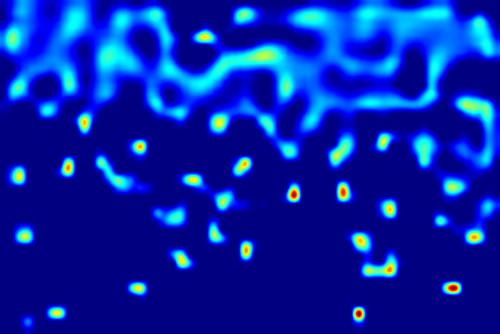} &
		\includegraphics[width=0.195\linewidth,height=0.15\linewidth]{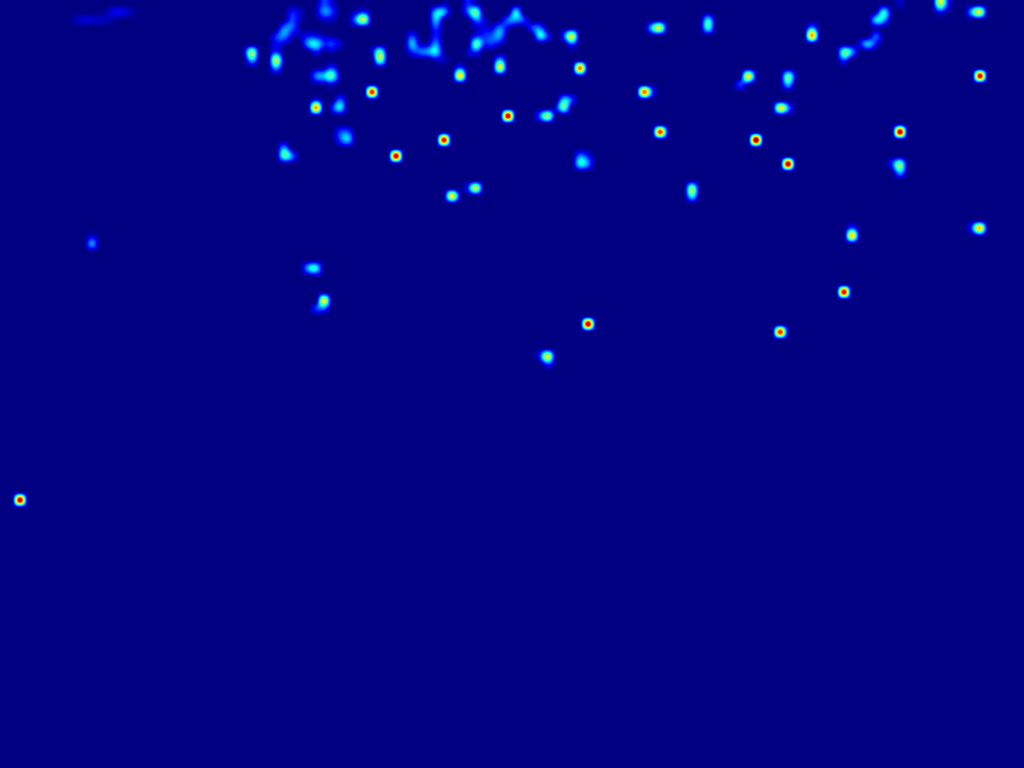} \\
		\small{Pred: 1160.6} & \small{Pred: 388.6} & \small{Pred: 253.1} & \small{Pred: 175.4} & \small{Pred: 83.9}\\
		\includegraphics[width=0.195\linewidth,height=0.15\linewidth]{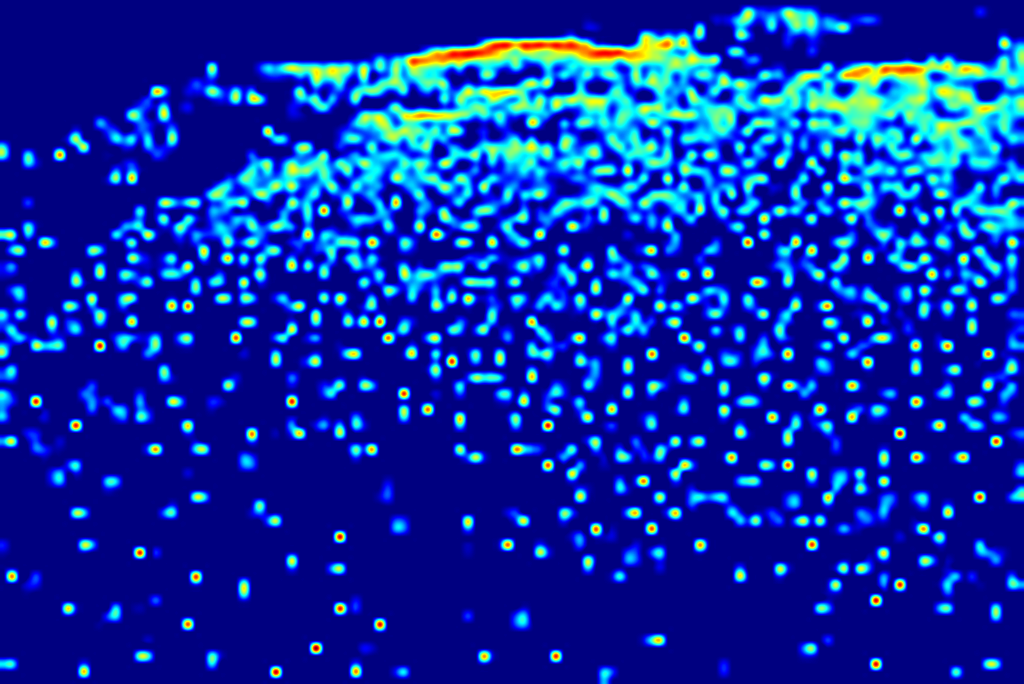} &
		\includegraphics[width=0.195\linewidth,height=0.15\linewidth]{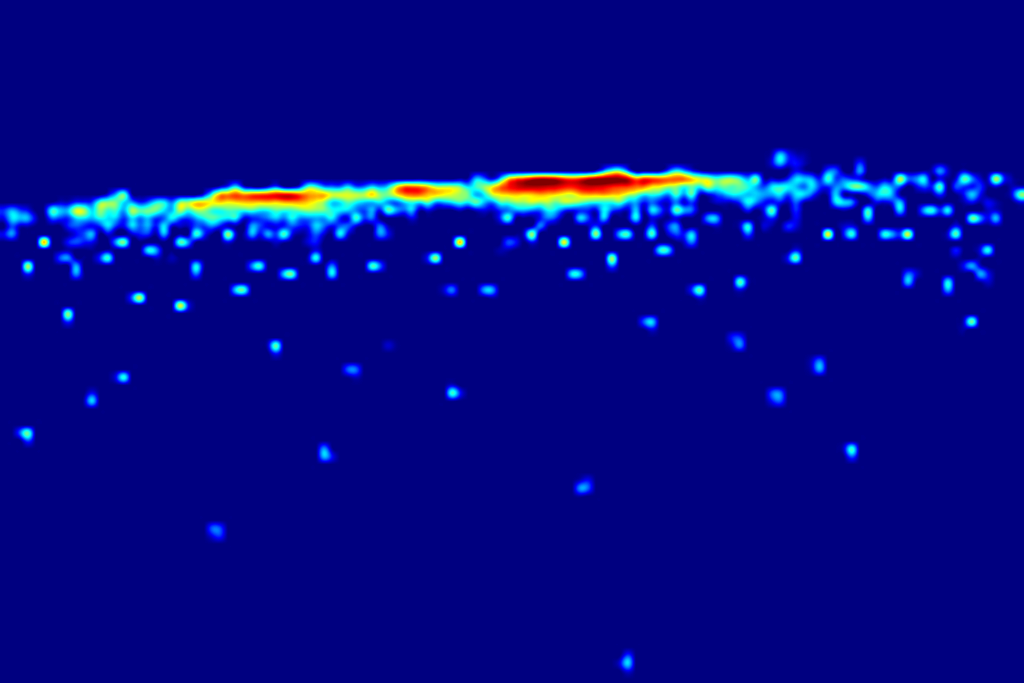} &
		\includegraphics[width=0.195\linewidth,height=0.15\linewidth]{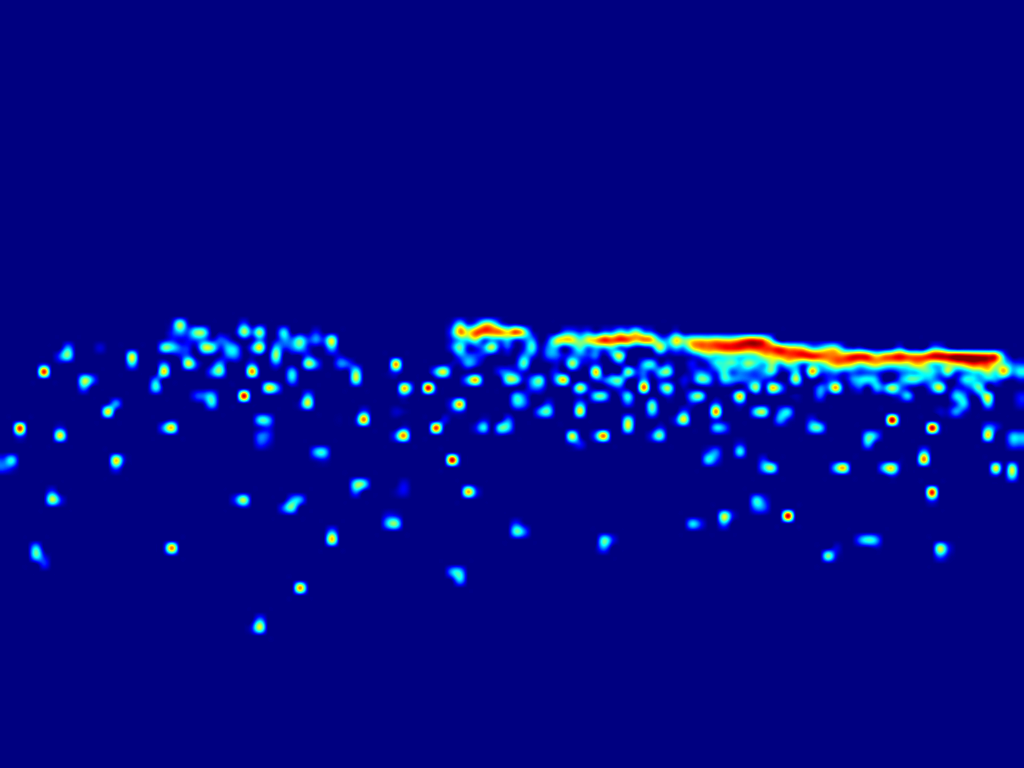} &
		\includegraphics[width=0.195\linewidth,height=0.15\linewidth]{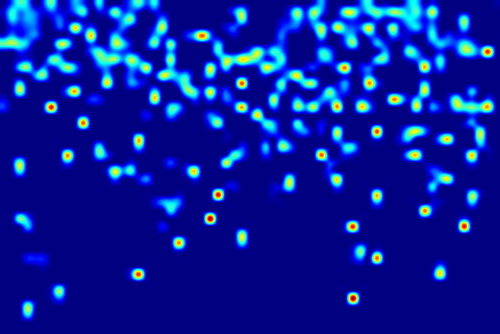} &
		\includegraphics[width=0.195\linewidth,height=0.15\linewidth]{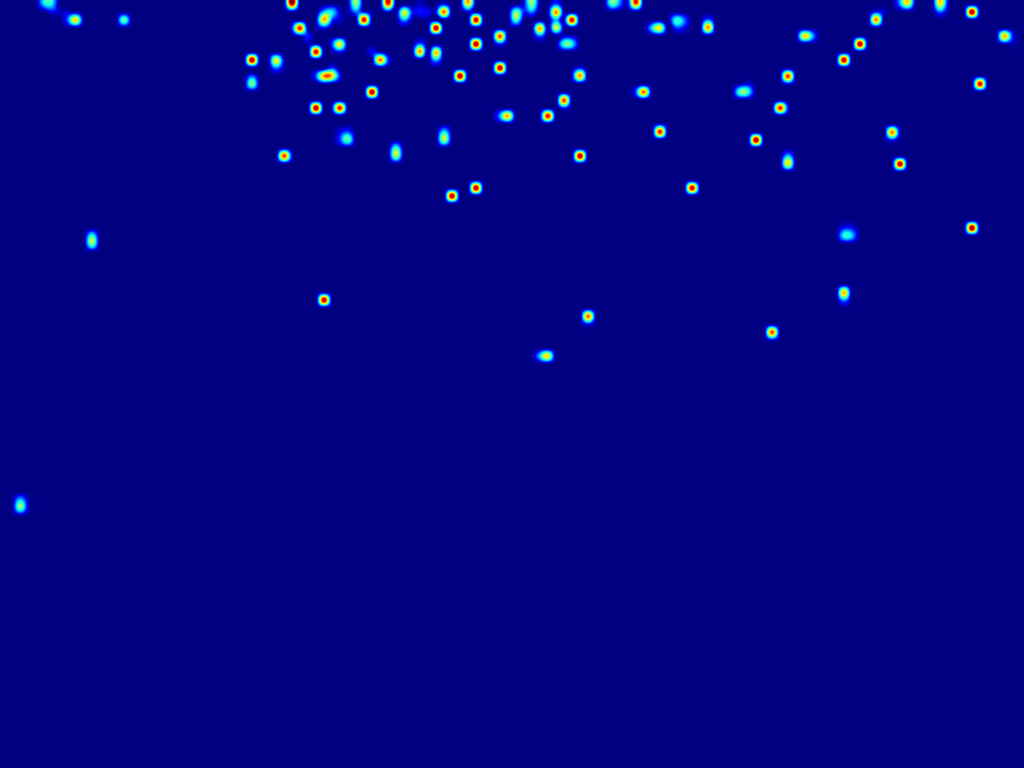} \\
	\end{tabular}
	\vspace{1mm}
	\caption{\textbf{Visualization of estimated density maps.} The first row: input images. The second row: estimated density maps of DM-Count for comparison. The third row: estimated density maps of our SAANet. GT: ground truth count, Pred: predicted count.
	}
	\label{fig:visdensitymap}
\end{figure*}

\subsection{Implement Details}
To implement SAANet, we first pre-train the proposed Deformer architecture on the image classification task. Then, we employ the pre-trained Deformer as the backbone network and train SAANet on crowd counting datasets.

\subsubsection{Pre-training of Deformer}
We pre-train the Deformer backbone on ImageNet-1K~\cite{deng2009imagenet}, which contains 1.28M images of 1000 categories. 
We train the model using the AdamW optimizer for 300 epochs with a cosine decay learning rate
scheduler and 20 epochs of linear warm-up.
A batch size of 1024, an initial learning rate of 0.001, and a weight decay of 0.05 are used.
We set the drop path rate to 0.2, 0.3, 0.5 for Deformer-tiny, Deformer-small, and Deformer-base.
The models are trained on 8 NVIDIA Tesla V100 GPUs.
We follow~\cite{wang2021pyramid,Liu_2021_ICCV,chu2021twins} for other hyper-parameters and training details.

\subsubsection{Crowd counting}
We adopt the pre-trained Deformer as the backbone and initialize the remaining modules randomly. We optimize the model with AdamW,  which is commonly adopted for transformer-based networks. The batch size is set to 8 for ShanghaiTech, and 16 for the other large-scale datasets. Other hyper-parameter settings and training strategies strictly follows~\cite{ma2019bayesian,wang2020distribution} to make a fair comparison.

\subsection{Comparison with State of The Arts}
\subsubsection{Benchmark results}
We compare our method with the current state-of-the-art methods on crowd counting benchmarks.
The results are shown in \Table{counting_results}, \Table{nwpu_result} and \Table{jhu_result}, respectively.
As can be seen, our method shows significant improvements over the other state of the arts.
SAANet surpasses both classical CNN-based methods and Transformer-based methods except for MFDC on UCF-QNRF, which adopts multiple experts of FDC with additional scene classification.
Compared with the contemporary CCTrans~\cite{tian2021cctrans} method, SAANet reduces the MAE by 4.3 and 3.3 on NWPU-Crowd val and test set, respectively.
The advantages are much more significant when compared with CNN-based methods, where SAANet reduces the MAE and MSE up to 36.2 and 277.8 on NWPU-Crowd val set as shown in~\Table{nwpu_result}.
We currently rank No.1 on the public leaderboard of NWPU-Crowd benchmark\footnote{https://www.crowdbenchmark.com/nwpucrowd.html}.
On JHU-Crowd++, our method improves the MAE across all the sub-categories and overall dataset consistently, as shown in~\Table{jhu_result}.
It can be seen that on large data sets, such as UCF-QNRF, NWPU-Crowd, JHU-Crowd++, the performance of SAANet is excellent due to its high capacity.

\subsubsection{Visualization of density maps}
We visualize the density maps predicted by our method and DM-Count~\cite{wang2020distribution}
for comparison. 
The two methods adopt the same loss function but different frameworks, where our method is a Transformer-based framework while DM-Count is based on CNN.
As shown in~\Fig{fig:visdensitymap}, the density maps estimated by our method appear sharp and are close to the locations of crowds.

\begin{table}[!t]
	\centering
	\caption{Efficiency and accuracy comparison of Deformer and other backbone networks.}
	\label{tab:backbone_results}
	\setlength{\tabcolsep}{4pt}
	\def\arraystretch{1.1}
	\resizebox{\linewidth}{!}{
		\begin{tabular}{l|c|c|cc|cc}
			\toprule
			\multirow{2}{*}{Backbone} & \multirow{2}{*}{FLOPs (G)} & \multirow{2}{*}{\#Param (M)} & \multicolumn{2}{c|}{SHA} & \multicolumn{2}{c}{NWPU (val)} \\
			\cmidrule{4-7}
			& & & MAE & MSE & MAE & MSE \\
			\midrule
			T2T-ViT-14~\cite{Yuan_2021_ICCV}  & 4.8 & 21.4  & 57.3 & 89.0 & 73.2 & 208.0 \\
			Swin-tiny~\cite{Liu_2021_ICCV}    & 7.3 & 27.5  & 58.6 & 89.7 & 45.8 & 114.5 \\ 
			Twins-small~\cite{chu2021twins}   & 4.3 & 23.6  & 60.6 & 94.3 & 43.7 & 109.2  \\ 
			Deformer-tiny                     & 4.3 & 27.6  & \textbf{53.9} & \textbf{88.7} & \textbf{41.2} & \textbf{93.9} \\
			\midrule
			T2T-ViT-19~\cite{Yuan_2021_ICCV}  & 8.5 & 39.2  & 60.7 & 94.1 & 59.5 & 181.3 \\
			Swin-small~\cite{Liu_2021_ICCV}   & 8.7 & 48.8  & 59.4 & 90.6 & 42.0 & 107.7 \\ 
			Twins-base~\cite{chu2021twins}    & 8.3 & 55.3  & 54.7 & 88.4 & 42.6 & 94.8  \\
			Deformer-small                    & 8.3 & 47.3  & \textbf{53.6} & \textbf{85.5} & \textbf{38.3} & \textbf{91.9} \\
			\midrule
			VGG-19~\cite{vgg}                 & 19.5 & 144.0  & 59.7 & 95.7 & 70.5 & 357.6 \\ 
			T2T-ViT-24~\cite{Yuan_2021_ICCV}  & 13.8 & 64.1  & 59.5 & 92.2 & 53.3 & 135.6 \\ 
			Swin-base~\cite{Liu_2021_ICCV}    & 25.2 & 86.7  & 57.2 & 88.9 & 46.6 & 111.5 \\ 
			Twins-large~\cite{chu2021twins}   & 21.7 & 98.3  & 57.9 & 94.2 & 41.1 & 95.6 \\ 
			Deformer-base                     & 19.4 & 84.1  & \textbf{53.8} & \textbf{86.3} & \textbf{37.5} & \textbf{83.9} \\
			\bottomrule
		\end{tabular}
	}
\end{table}

\subsection{Ablation Study}
As aforementioned, our proposed SAANet is composed of the following components: a Deformer backbone to extract features, a multi-level feature fusion module to perform multi-scale information fusion, and a count-attentive feature enhancement module to strengthen feature representations.
We conduct a number of ablation experiments to verify the effectiveness of each component in SAANet.

\begin{figure*}[!tb]\small
	\centering
	\begin{minipage}{0.53\linewidth}
		\setlength{\tabcolsep}{1pt}
		\def\arraystretch{0.8}
		\begin{tabular}{cc}
			\includegraphics[width=0.49\linewidth,height=0.33\linewidth]{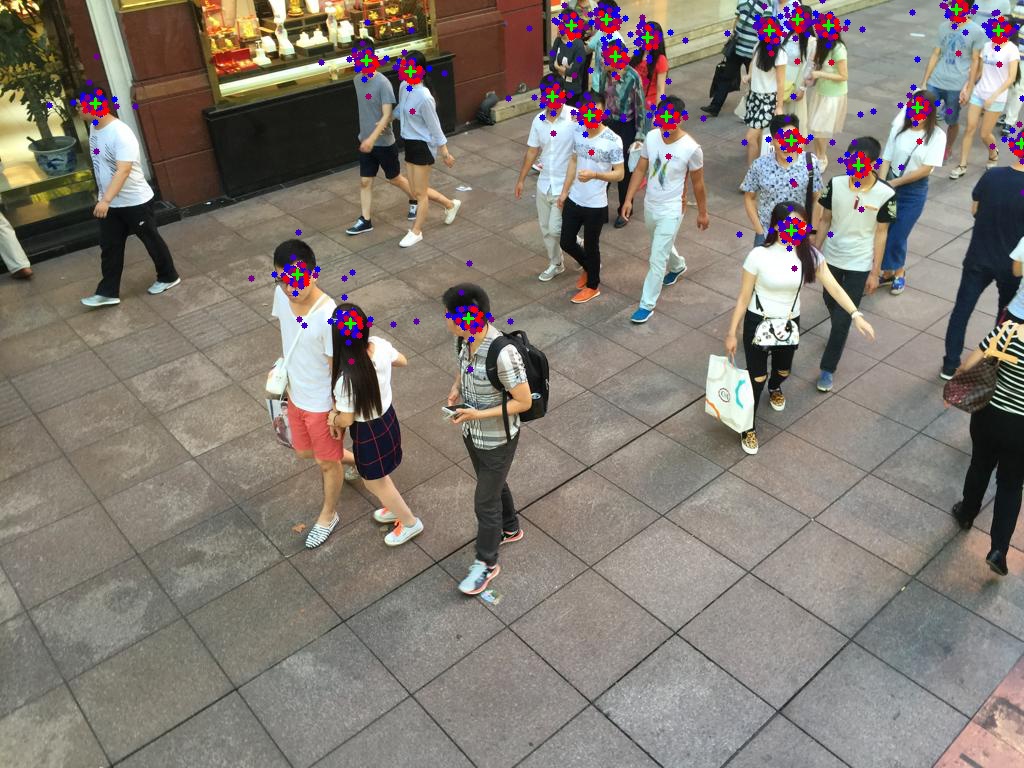} & 
			\includegraphics[width=0.49\linewidth,height=0.33\linewidth]{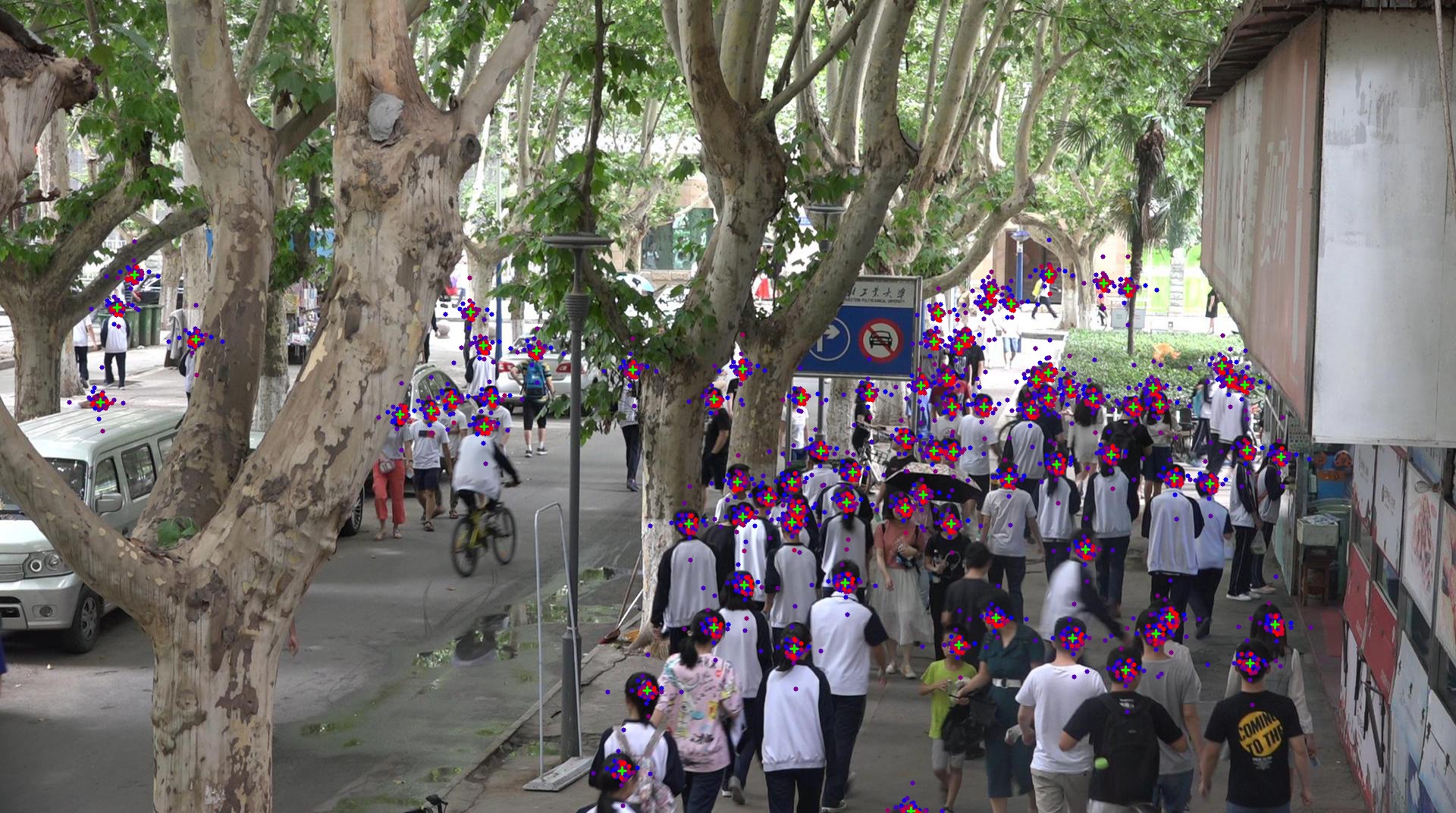} \\
			\includegraphics[width=0.49\linewidth,height=0.33\linewidth]{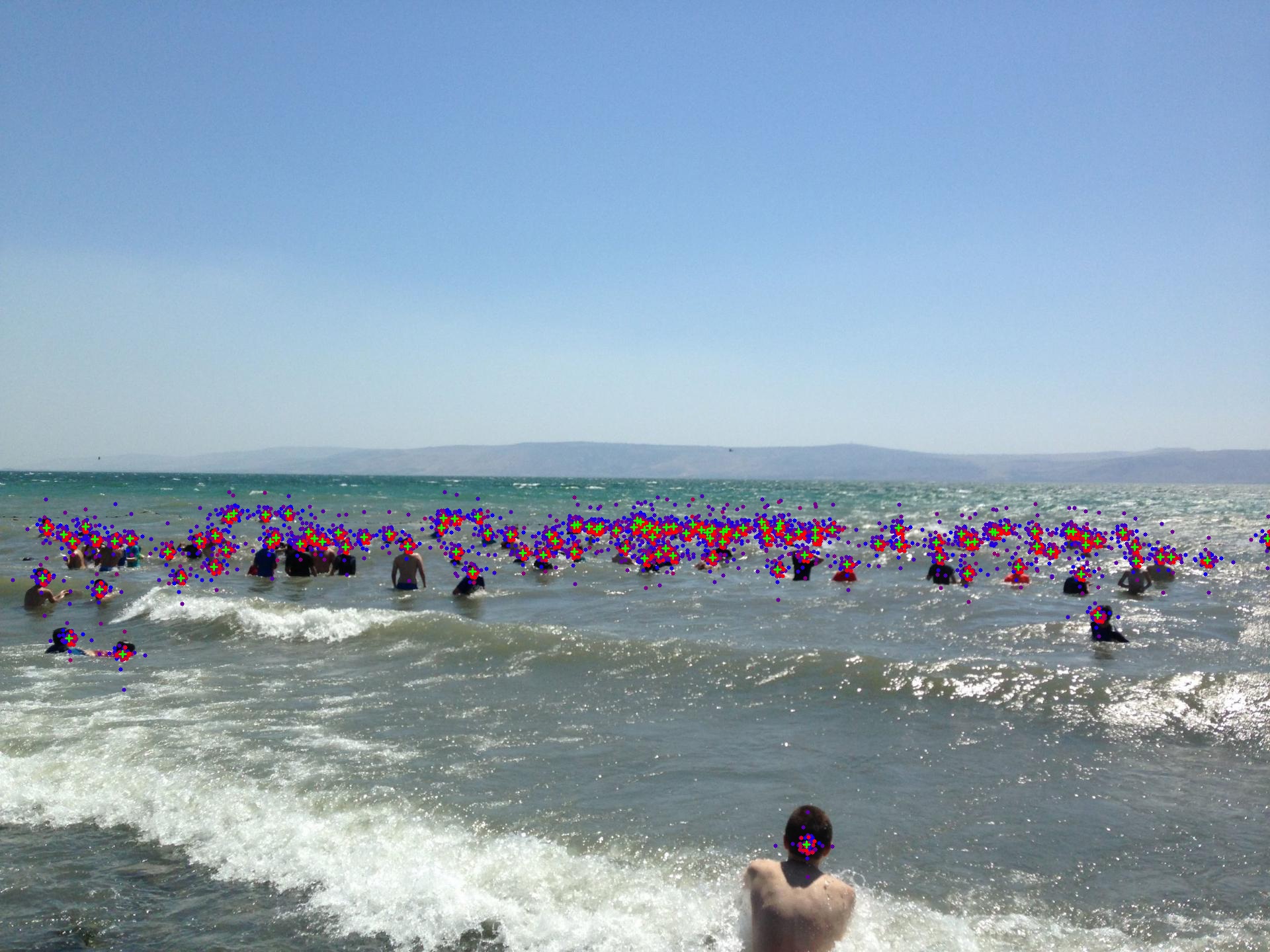} & 
			\includegraphics[width=0.49\linewidth,height=0.33\linewidth]{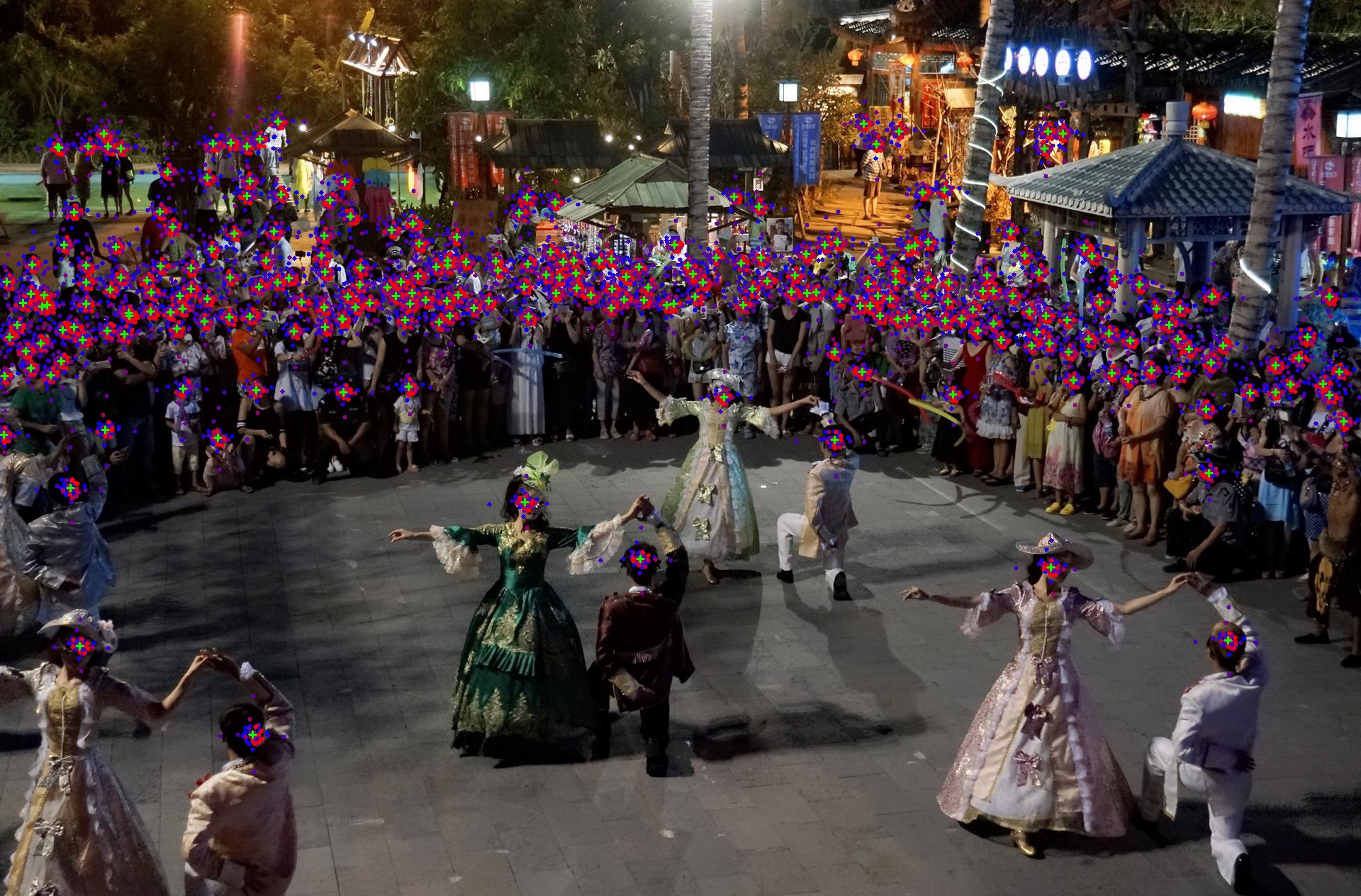}\\
		\end{tabular}
	\end{minipage}
	\hfill
	\begin{minipage}{0.44\linewidth}
		\begin{tikzpicture}[/pgfplots/width=1.0\linewidth,/pgfplots/height=0.92\linewidth]
		\begin{axis}[ymin=18,ymax=46,xmin=0,xmax=400,
		xlabel={Size of Bounding-box},
		ylabel={Sampling Offset},
		yticklabel style={/pgf/number format/fixed,/pgf/number format/precision=2},
		xtick={0,50,100,150,200,250,300,350,400},
		font=\scriptsize,
		grid=both,
		grid style=dotted,
		major grid style={white!20!black},
		minor grid style={white!70!black},
		axis equal image=false]
		\addplot[red,smooth,ultra thick,domain=0:400]{0.03675*x+28.857};
		\addplot+[green,mark options={fill=blue},only marks,mark=*,mark size=0.8pt] table[x=x,y=y] {./fig/fig6/scale-sampling-offset.txt};
		\end{axis}
		\end{tikzpicture}
	\end{minipage}
	\caption{
		\textbf{Visualization and analysis of the sampling offsets in Deformer.} Left: visualization of estimated sampling points. The ground-truth points are denoted as a green cross marker. And each estimated sampling point is marked as a filled circle whose color indicates its corresponding attention weight. The warmer color, the higher the attention weight. Right: the correlation between learned sampling offsets and the size of bounding-boxes on JHU-Crowd++.}
	\label{fig:samplingpoints}
\end{figure*}
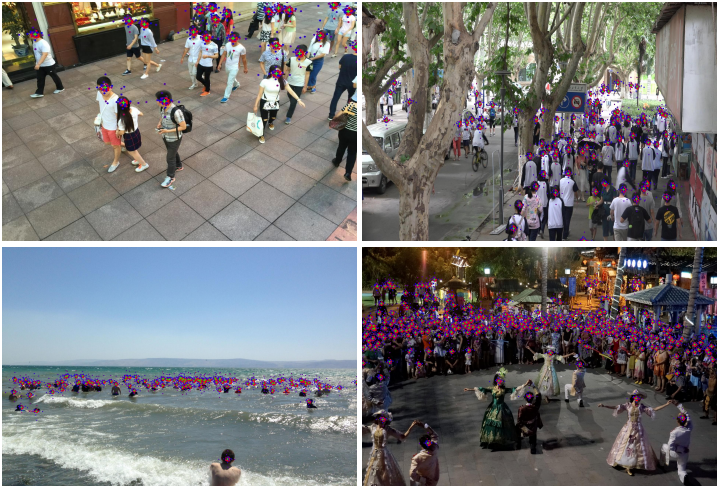

\subsubsection{Advantages of the Deformer backbone}
\label{sec:ablation-deformer-backbone}
To show the advantages of the proposed Deformer backbone, we compare the Deformer series with other state-of-the-art CNN- and transformer-based backbones.
We use the same prediction head (without involving other proposed modules) and loss function~\cite{wang2020distribution} for all backbones to ensure a fair comparison.
In addition to counting accuracy, we also compare the computational complexity and parameters consumed by each backbone.
The FLOPs of each compared model are calculated on inputs of the same $224\times224$ resolution.
As can be seen from \Table{backbone_results}, our method outperforms others significantly using fewer model parameters and comparable computational costs.
Moreover, our lightweight model Deformer-small even achieves better performance than the larger model Twins-large and Twins-base, demonstrating our proposed backbone's high efficiency and applicability.

~\Fig{fig:samplingpoints} shows the visualization of the sampling points of the deformable attention in our Deformer backbone.
We visualize the estimated sampling points in the second stage of Deformer ($\frac{1}{8}$ of the image resolution).
Moreover, we analyze the distribution of learned sampling offsets with respect to the scale of bounding boxes on JHU-Crowd++. 
As shown in the right of~\Fig{fig:samplingpoints}, the sampling offsets are positively correlated with the size of bounding-boxes.
Note that the bounding-box information is not used in training.
These reveal that the deformable attention module can adapt its sampling locations and the corresponding attention weights according to the different scales of human heads.

\subsubsection{Ablation on multi-level feature fusion}
To show the effect of the MFF module, we first perform a baseline experiment where the single-level (final layer) feature is used to predict the density map directly.
We then enable the MFF module to fuse the multi-level features with deformable attentions.
As is shown in the~\Table{decoder_results}, the feature fusion module can reduce MAE/MSE from 53.8/86.3 to 53.2/84.2, compared to the baseline.
Moreover, we also perform an ablation study using the VGG backbone.
The MFF module consistently improves the performance with different backbones.

\subsubsection{Ablation on count-attentive feature enhancement}
We further perform an ablation study to investigate the effect of the CAFE module.
The CAFE module is a stack of transformer decoder layers with count loss supervision, where the multi-head attention maps are used to re-calibrate the feature maps.
As shown in \Table{decoder_results}, in comparison with the baseline model, the proposed CAFE module can reduce MAE/MSE from 53.8/86.3 to 52.8/83.2.
After that, we conduct another experiment to enable both MFF and CAFE modules simultaneously.
It can lead to a noticeable improvement, reducing MAE and MSE by 2.1 and 6.2, respectively.
Besides, the same improvement can be observed using the VGG backbone, further demonstrating the effectiveness and generalization of the proposed MFF and CAFE modules.

To visualize the effect of CAFE, we accumulate attention maps across all heads of the last cross attention layer and then draw the attention maps. As shown in~\Fig{fig:attention}, the lighter region means where the decoder pays more \emph{attention} on.
As can be seen, the attention maps could capture high-level semantic information such that they tend to focus on the foreground regions.

\begin{table}[t]
	\centering
	\caption{Ablation study. MFF: Multi-level Feature Fusion, CAFE: Count-Attentive Feature Enhancement.}
	\label{tab:decoder_results}
	\setlength{\tabcolsep}{10pt}
	\def\arraystretch{1.1}
	\begin{tabular}{l|cc|cc}
		\toprule
		\multirow{2}{*}{Backbone} & \multirow{2}{*}{MFF} & \multirow{2}{*}{CAFE} &\multicolumn{2}{c}{SHA} \\
		\cmidrule{4-5}
		& & & MAE & MSE \\
		\midrule
		\multirow{4}{*}{Deformer-base}  & & & 53.8 & 86.3 \\
		& $\checkmark$ & & 53.2 & 84.2 \\
		& & $\checkmark$ &52.8 &83.2 \\
		& $\checkmark$ & $\checkmark$ & \textbf{51.7} & \textbf{80.1} \\
		\midrule
		\multirow{4}{*}{VGG-19} & & & 59.7 & 95.7 \\
		& $\checkmark$ & & 58.5 & 93.5 \\
		& & $\checkmark$ & 57.3 & 95.3 \\
		& $\checkmark$ & $\checkmark$ & \textbf{56.4} & \textbf{91.6}\\
		\bottomrule
	\end{tabular}
\end{table}

\begin{figure*}[t]\small
	\centering
	\setlength{\tabcolsep}{1pt}
	\begin{tabular}{ccccc}
		\includegraphics[width=0.195\linewidth,height=0.16\linewidth]{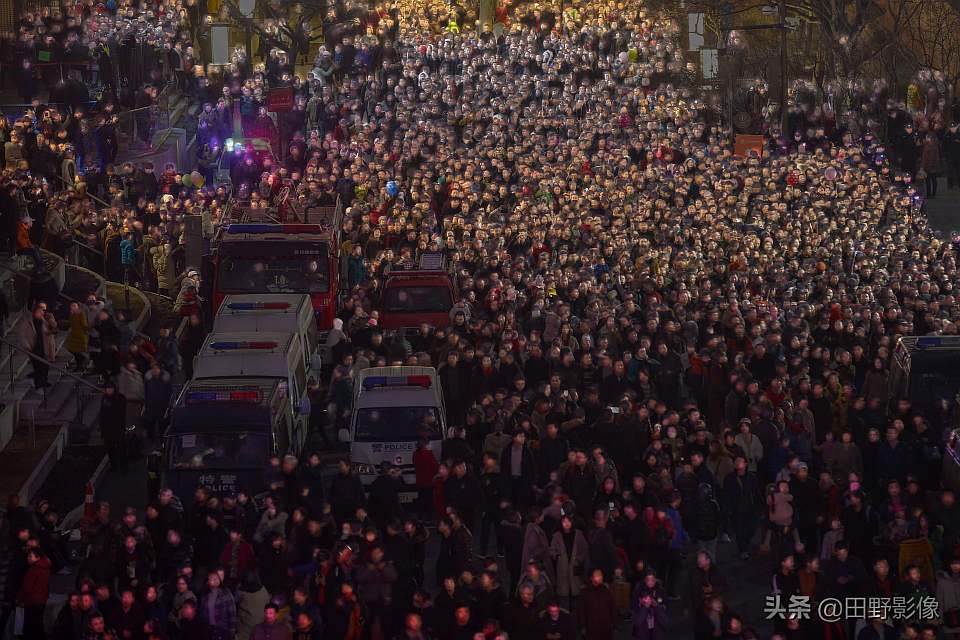} &
		\includegraphics[width=0.195\linewidth,height=0.16\linewidth]{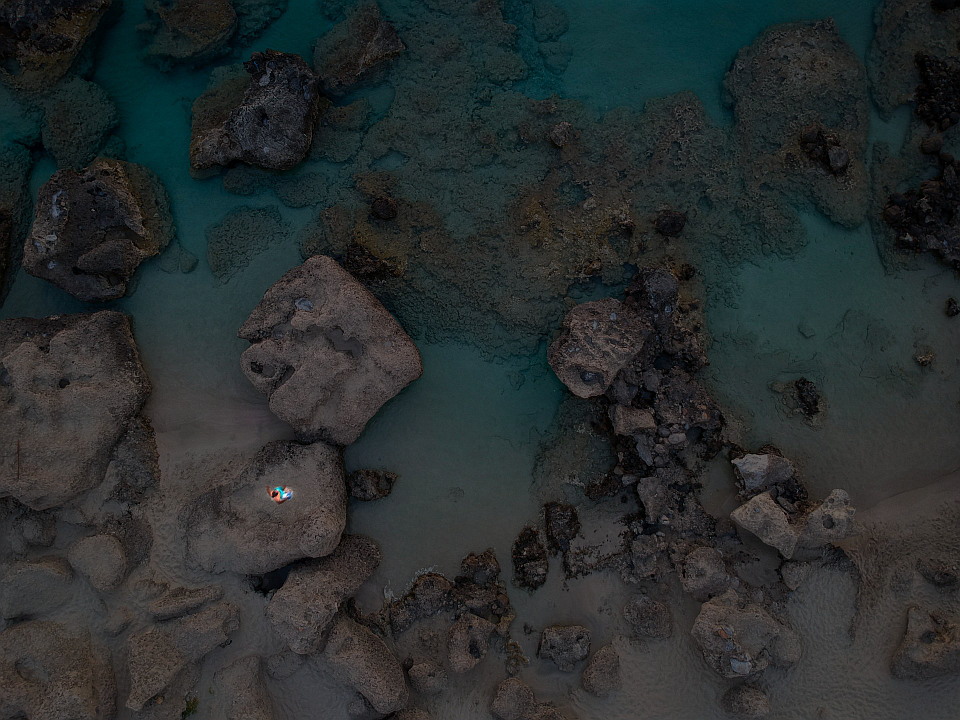} &
		\includegraphics[width=0.195\linewidth,height=0.16\linewidth]{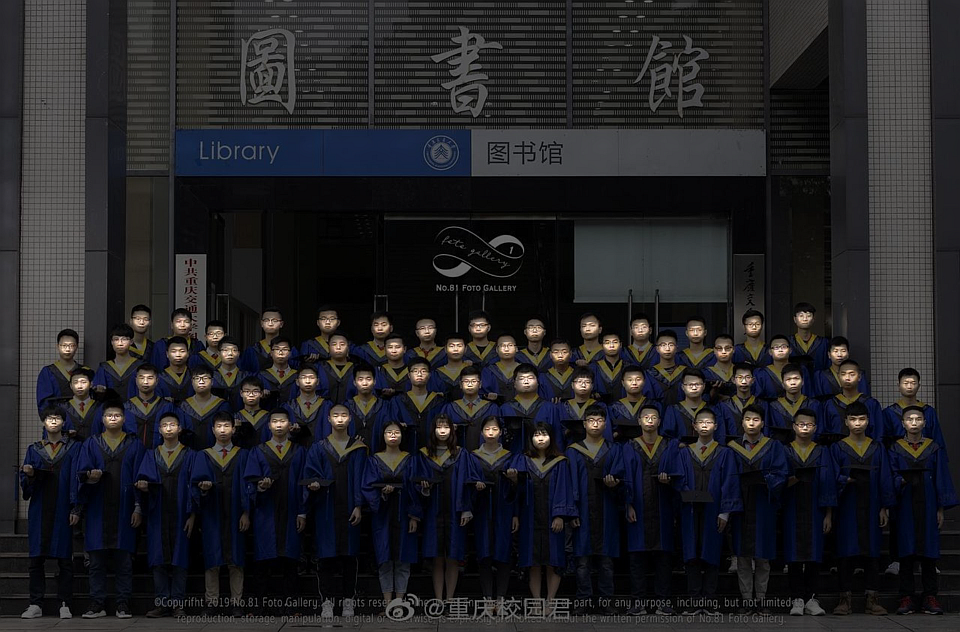} & 
		\includegraphics[width=0.195\linewidth,height=0.16\linewidth]{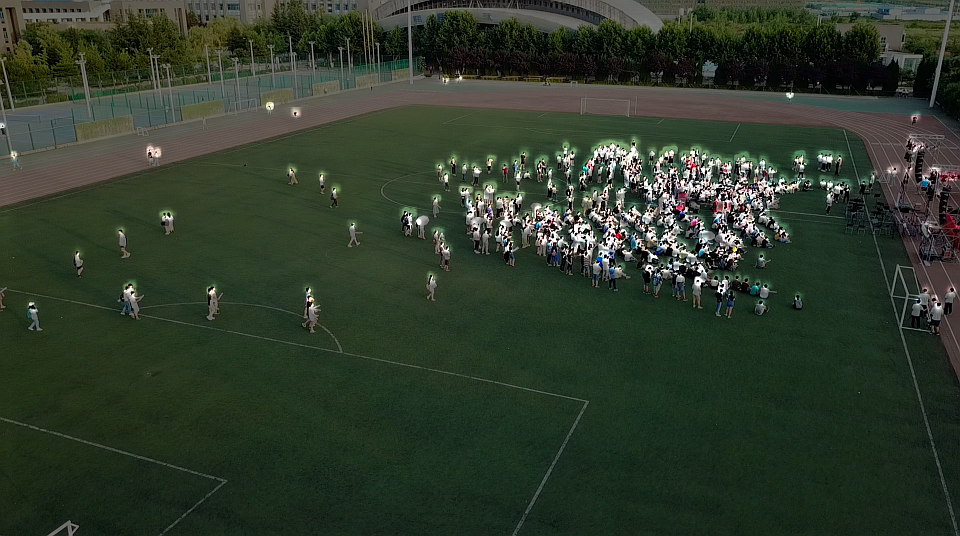} &
		\includegraphics[width=0.195\linewidth,height=0.16\linewidth]{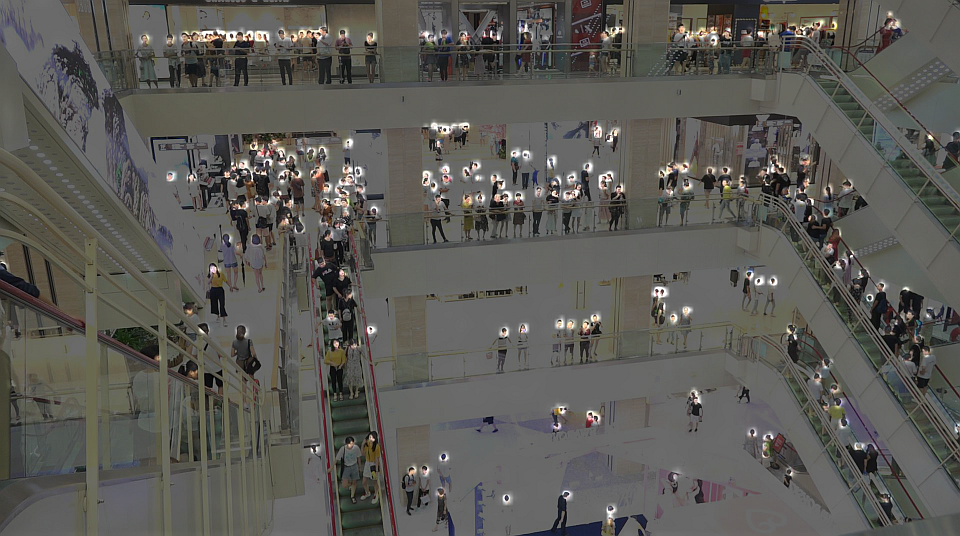}
		\\
		\includegraphics[width=0.195\linewidth,height=0.16\linewidth]{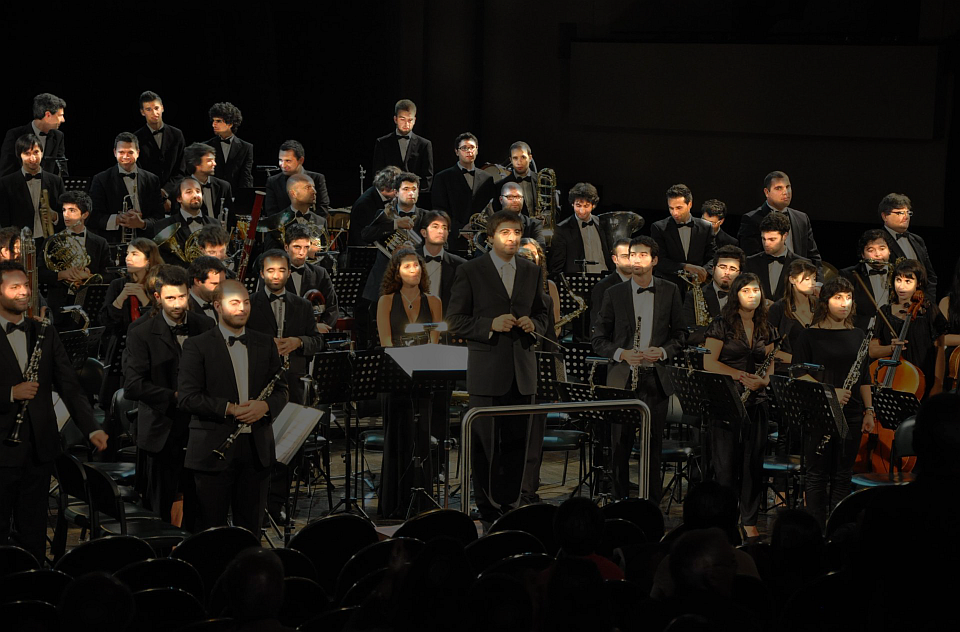} &
		\includegraphics[width=0.195\linewidth,height=0.16\linewidth]{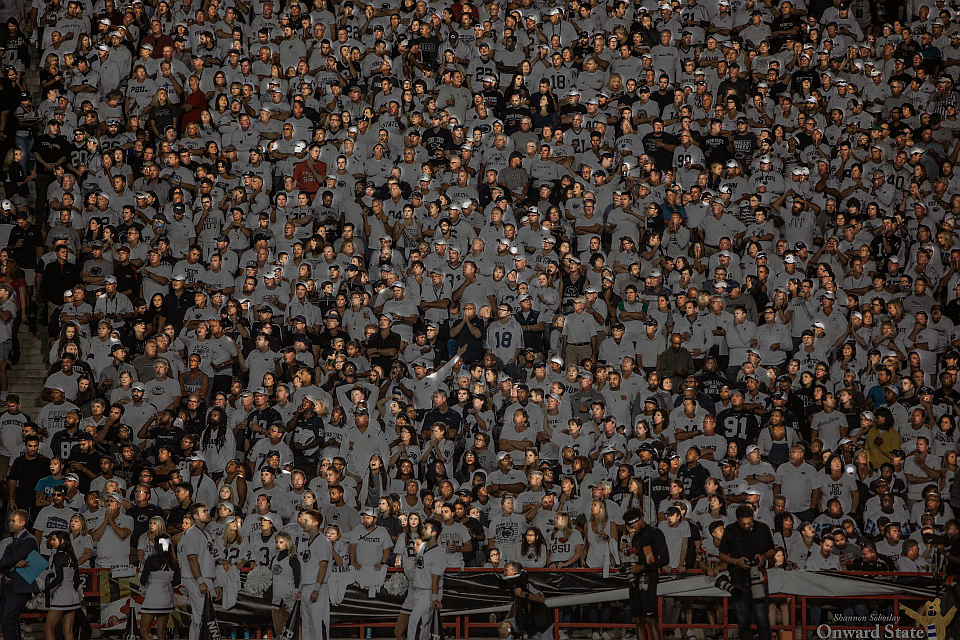} &
		\includegraphics[width=0.195\linewidth,height=0.16\linewidth]{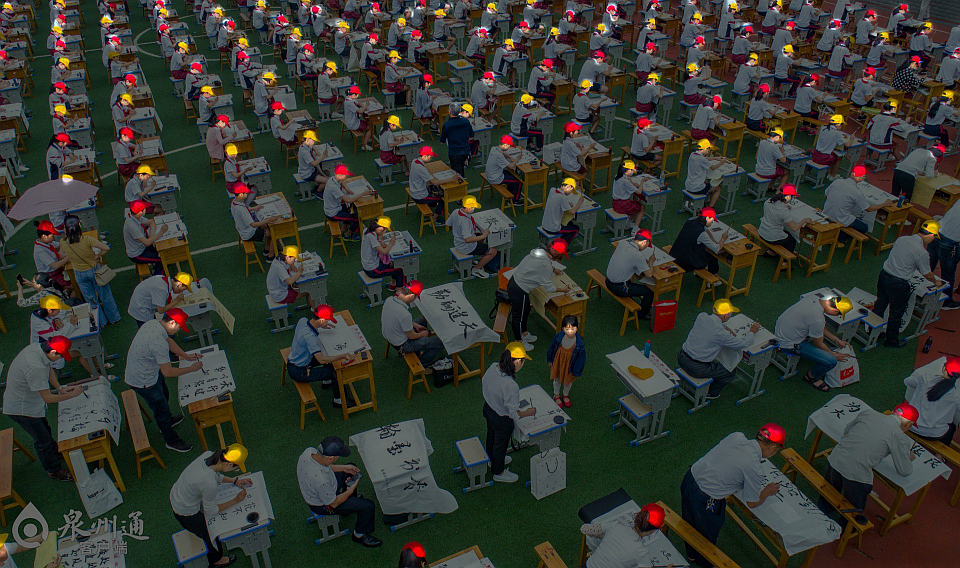} & 
		\includegraphics[width=0.195\linewidth,height=0.16\linewidth]{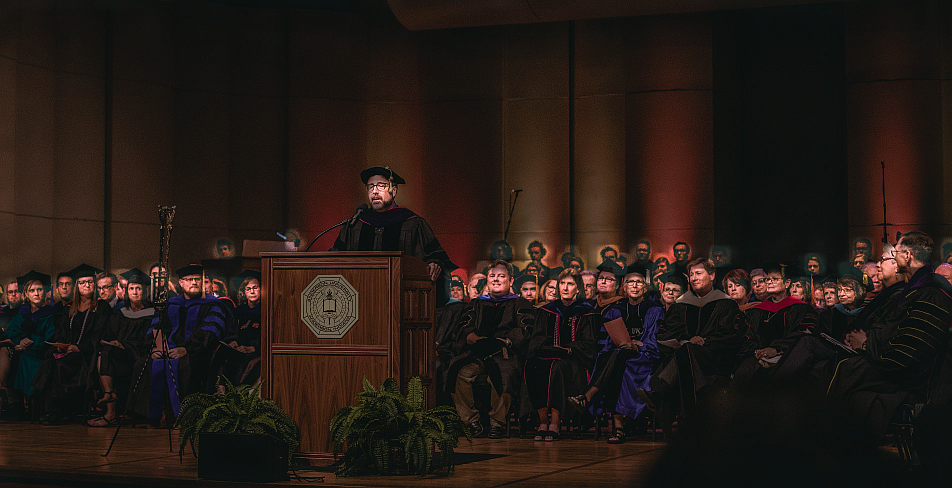} &
		\includegraphics[width=0.195\linewidth,height=0.16\linewidth]{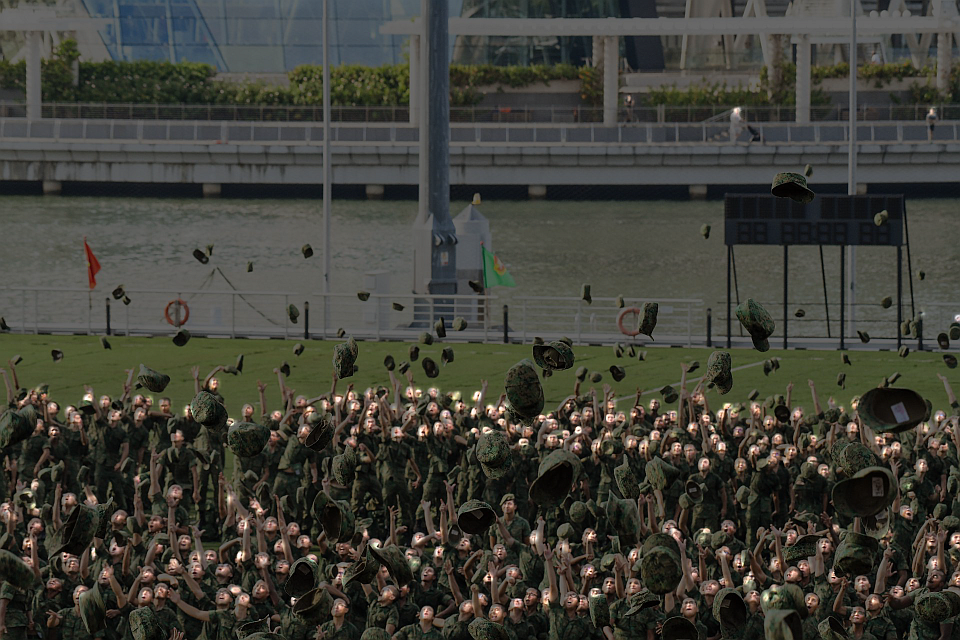}
	\end{tabular}
	\caption{\textbf{Visualization of attention maps in the CAFE module.} We visualize the attention maps of the last decoder layer, which are used to re-calibrate the feature maps produced by the preceding DFF module. The lighter region means where the decoder pays more attention on.}
	\label{fig:attention}
\end{figure*}

%% file: sec/5_conclusion.tex
\section{Conclusions}

In this paper, we propose a transformer-based framework for crowd counting. 
Our method, termed SAANet, is designed to avoid the limitations of convolution networks and recent Transformer architectures with windowed attention which are not adaptive to the scene.
For this purpose, we propose a deformable attention in-built architecture for crowd counting.
Our method learns adaptive feature representations with deformable sampling locations and dynamical attention weights, which increase the ability to handle scale and scene variations.
Besides, we propose two modules to enhance feature representation further.
The experiment results show that our method plays favorably against state-of-the-art approaches on public crowd counting benchmarks.